\PassOptionsToPackage{unicode}{hyperref}
\PassOptionsToPackage{hyphens}{url}
\documentclass[
  american,
  a4paper,
  10pt]{article}
\usepackage{lmodern}
\usepackage{setspace}
\usepackage{amssymb,amsmath}
\usepackage{ifxetex,ifluatex}
\ifnum 0\ifxetex 1\fi\ifluatex 1\fi=0 
  \usepackage[T1]{fontenc}
  \usepackage[utf8]{inputenc}
  \usepackage{textcomp} 
\else 
  \usepackage{unicode-math}
  \defaultfontfeatures{Scale=MatchLowercase}
  \defaultfontfeatures[\rmfamily]{Ligatures=TeX,Scale=1}
\fi
\IfFileExists{upquote.sty}{\usepackage{upquote}}{}
\IfFileExists{microtype.sty}{
  \usepackage[]{microtype}
  \UseMicrotypeSet[protrusion]{basicmath} 
}{}
\makeatletter
\@ifundefined{KOMAClassName}{
  \IfFileExists{parskip.sty}{%
    \usepackage{parskip}
  }{
    \setlength{\parindent}{0pt}
    \setlength{\parskip}{6pt plus 2pt minus 1pt}}
}{
  \KOMAoptions{parskip=half}}
\makeatother
\usepackage{xcolor}
\IfFileExists{xurl.sty}{\usepackage{xurl}}{} 
\IfFileExists{bookmark.sty}{\usepackage{bookmark}}{\usepackage{hyperref}}
\hypersetup{
  pdftitle={CubicQuant: Parametric Non-Uniform Codebooks for High-Throughput LLM Inference with 1--8-Bit Weights},
  pdfauthor={Xuetian Gao (Elliot Gao)},
  pdflang={en-US},
  hidelinks,
  pdfcreator={LaTeX via pandoc}}
\usepackage[margin=25mm]{geometry}
\usepackage{color}
\usepackage{fancyvrb}

\DefineVerbatimEnvironment{Highlighting}{Verbatim}{commandchars=\\\{\}}
\newenvironment{Shaded}{}{}

\newcommand{\BuiltInTok}[1]{#1}

\newcommand{\ControlFlowTok}[1]{\textcolor[rgb]{0.00,0.44,0.13}{\textbf{#1}}}

\newcommand{\DecValTok}[1]{\textcolor[rgb]{0.25,0.63,0.44}{#1}}

\newcommand{\FloatTok}[1]{\textcolor[rgb]{0.25,0.63,0.44}{#1}}

\newcommand{\KeywordTok}[1]{\textcolor[rgb]{0.00,0.44,0.13}{\textbf{#1}}}
\newcommand{\NormalTok}[1]{#1}
\newcommand{\OperatorTok}[1]{\textcolor[rgb]{0.40,0.40,0.40}{#1}}

\usepackage{longtable,booktabs}
\usepackage{etoolbox}
\makeatletter
\patchcmd\longtable{\par}{\if@noskipsec\mbox{}\fi\par}{}{}
\makeatother
\IfFileExists{footnotehyper.sty}{\usepackage{footnotehyper}}{\usepackage{footnote}}
\makesavenoteenv{longtable}
\providecommand{\tightlist}{%
  \setlength{\itemsep}{0pt}\setlength{\parskip}{0pt}}
\usepackage{microtype}
\usepackage{amssymb}
\usepackage{booktabs}
\usepackage{array}
\usepackage{tabularx}
\usepackage{siunitx}
\usepackage{enumitem}
\usepackage{pgfplots}
\pgfplotsset{compat=1.18}
\usepackage[font=small,labelfont=bf]{caption}
\usepackage[nameinlink,noabbrev]{cleveref}

\setlist{topsep=0.25em,itemsep=0.12em,parsep=0pt,leftmargin=*}
\AtBeginDocument{%
  \hypersetup{
    pdfborder={0 0 0}
  }%
}
\ifxetex
  \usepackage{polyglossia}
  \setmainlanguage[variant=american]{english}
\else
  \usepackage[shorthands=off,main=american]{babel}
\fi

\title{CubicQuant: Parametric Non-Uniform Codebooks for High-Throughput
LLM Inference with 1--8-Bit Weights}
\author{Xuetian ``Elliot'' Gao\\
\small \href{mailto:gxtcch@outlook.com}{\nolinkurl{gxtcch@outlook.com}}}
\date{Technical Report \textperiodcentered{} August 2026}

\begin{document}
\maketitle

\setstretch{1.04}
\begin{abstract}

Weight quantization for large-language-model inference must reconcile two
competing goals: reconstruction levels should adapt to the strongly varying
local statistics of model weights, yet their representation should remain
compact and regular enough for direct GPU execution. Uniform integers favor
regularity but constrain every group to a linear grid. Low-bit floating-point
formats redistribute levels through a fixed exponent--mantissa structure,
whereas learned codebooks gain flexibility at the cost of less regular
decoding and additional metadata.

We introduce CubicQuant, a parametric non-uniform scalar format that preserves a
dense integer code stream while adapting its reconstruction levels independently
within each weight group. A monotonic cubic curve, specified by two shape
parameters and one scale, maps uniformly spaced magnitude codes to non-uniform
levels. The family spans 1--8-bit weight payloads, contains symmetric uniform
integer quantization as an exact special case, and has effective width
$B+64/G$ bits per weight for payload width $B$ and group size $G$. We derive
its population distortion under Uniform, Gaussian, and Laplace distributions,
formulate both continuous and Dynamic-A8-carrier-aware fitting objectives, and
describe direct packed-weight GPU execution with model-dtype or dynamically
quantized INT8 activations.

In a finite-group G128 experiment with 15,360 samples per distribution, W4
CubicQuant reduced reconstruction RMSE relative to optimally clipped four-bit
uniform integer quantization by 3.90\% on Uniform, 13.49\% on Gaussian, and
28.14\% on Laplace samples. Relative to the best enumerated four-bit finite
floating-point format, the reductions were 3.90\%, 9.44\%, and 6.27\%.
Preliminary H200 kernel measurements further reveal a workload-dependent
crossover: model-dtype execution is faster for narrow GEMV, while Dynamic A8
becomes favorable as row count grows. Together, the results establish the
representational promise and executable character of the format, while leaving
downstream model quality and cross-device end-to-end performance as open
evaluation questions.
\end{abstract}

\hypertarget{introduction}{%
\section{Introduction}\label{introduction}}

Autoregressive inference repeatedly streams large weight matrices
through GPU memory. Reducing their stored width directly relieves
capacity and bandwidth pressure, particularly in decode-like regimes
where weight reuse across tokens is limited. At low precision, however,
storage width alone is not sufficient: the placement of the few
available reconstruction levels determines whether a format preserves
the mass near zero, the tails, or neither. A useful format must
therefore balance statistical adaptability against the regularity
required by high-throughput matrix kernels.

The established scalar choices occupy distinct points on this trade-off.
Uniform integers offer dense packing and simple arithmetic, but every
group is restricted to an affine level grid. Floating-point formats
devote bits to a fixed exponent--mantissa hierarchy and thereby favor a
predetermined dynamic range. Free or learned codebooks can adapt all
levels, yet their lookup and metadata structure is less naturally
aligned with conventional GPU dot-product paths. Vector and additive
codebooks enlarge the design space further, but also change the decoding
problem from scalar reconstruction to structured lookup.

CubicQuant occupies the middle ground between a fixed grid and a free
codebook. It retains scalar integer codes and a regular packed
bitstream, but makes the represented level positions parametric. Within
each group, a normalized monotonic cubic maps uniformly spaced magnitude
indices to non-uniform reconstruction levels. Two shape coefficients
control the distribution of interior levels and a scale fixes the
endpoint. The resulting codebook is adaptive, yet sufficiently
structured to be regenerated from compact metadata inside the consuming
GPU tile.

Four requirements shape the format:

\begin{enumerate}
\def\labelenumi{\arabic{enumi}.}
\tightlist
\item
  \textbf{Compactness.} Weight codes remain densely packed from one to
  eight bits, and metadata cost is explicit rather than hidden in a
  model-level average.
\item
  \textbf{A stable numerical contract.} Zero and both signed endpoints
  are exact; the scale is persisted and applied in FP32; the codebook is
  monotonic.
\item
  \textbf{Estimator independence.} The representation does not prescribe
  how its codes and parameters are estimated; data-free,
  activation-aware, and second-order objectives can share the same
  format.
\item
  \textbf{Direct GPU execution.} Cubic weights can be consumed directly
  by GPU kernels with either BF16/FP16 activations or dynamically
  quantized INT8 activations. The packed weights are reconstructed
  within each compute tile rather than materialized as a full-precision
  tensor, while the kernel strategy may vary with tensor shape and GPU
  architecture without changing the numerical format.
\end{enumerate}

This report develops that design through five contributions:

\begin{itemize}
\tightlist
\item
  a symmetric, groupwise, two-parameter family of scalar non-uniform
  codebooks that contains uniform signed integer quantization as an
  exact special case;
\item
  a packed checkpoint format spanning one- to eight-bit weights, with
  FP32 scaling and a closed-form storage overhead;
\item
  a groupwise reference fitting procedure and a Dynamic-A8-carrier-aware
  objective, both separable from richer data-driven error estimators;
\item
  GPU execution paths for model-dtype and Dynamic-A8 activations, with
  device- and tensor-shape-aware calibration; and
\item
  analytical memory and operation models together with preliminary
  numerical and Hopper kernel evidence that delimit the present claims.
\end{itemize}

CubicQuant is developed within the QuantTrio open-source project. Its
reference implementation uses vLLM as a systems substrate, while the
contribution studied here is confined to the weight representation,
fitting objectives, packed operators, and their evaluation. Scheduler,
attention, and memory-management behavior are not attributed to the
format \cite{kwon2023}.

The remainder of the report follows the format from definition to
evidence. Section 2 specifies the code space, level function, and
metadata. Section 3 derives population distortion and connects it to
finite-group behavior. Section 4 formulates offline fitting, including
the deterministic Dynamic-A8 carrier projection. Section 5 describes
packed GPU realization, and Section 6 separates analytical properties
from the presently available numerical and kernel evidence. Sections
7--9 position the work, state its limitations, and conclude.

\hypertarget{cubicquant-representation}{%
\section{CubicQuant representation}\label{cubicquant-representation}}

Throughout this report, \(W_bA_a\) denotes a \(b\)-bit weight payload
and an \(a\)-bit activation precision; W4A8 therefore means four-bit
weights with eight-bit activations. After this definition, W1--W8 is
used as a range notation for the eight supported weight payload widths,
not as the name of a single format.

\hypertarget{signed-codes}{%
\subsection{Signed codes}\label{signed-codes}}

Let \(B\) be the payload width. For \(B>1\), define

\begin{equation}
M = 2^{B-1}-1, \qquad
k \in \{-M,\ldots,0,\ldots,M\}.
\label{eq:codes}
\end{equation}

The signed code range has \(2^B-1\) valid values. The remaining
two's-complement bit pattern, \(-2^{B-1}\), is reserved and decodes
deterministically to zero. This choice gives an exact zero while
preserving symmetric positive and negative endpoints. W1 is a separate
binary case with codes in \(\{-1,+1\}\) and no zero code.

Packed codes are stored as a continuous bitstream in row-major logical
order. At widths that do not divide eight, such as W3, W5, W6, and W7,
an individual code may span two adjacent bytes. The format therefore
specifies both cross-byte decoding and the treatment of unused bits at
the end of each logical row.

\hypertarget{parametric-level-function}{%
\subsection{Parametric level function}\label{parametric-level-function}}

For \(B>1\), reconstruction is defined by

\begin{equation}
\begin{aligned}
t_k &= \frac{|k|}{M}, \\
c &= 1-a-b, \\
q(t) &= t\left[a+t\left(b+ct\right)\right], \\
\widehat{w}(k) &= \operatorname{sign}(k)\,s\,q(t_k).
\end{aligned}
\label{eq:cubic}
\end{equation}

The reconstruction combines a stored integer code with group-local scale
and shape parameters. The notation and the role of each quantity are
summarized below.

\begingroup
\renewcommand{\arraystretch}{1.12}
\noindent\begin{tabularx}{\linewidth}{@{}l>{\raggedright\arraybackslash}X@{}}
\toprule
\textbf{Symbol} & \textbf{Definition} \\
\midrule
$B$ & Weight payload width in bits. Equation \eqref{eq:cubic} applies to
$B\in\{2,\ldots,8\}$; W1 is the binary case defined separately above. \\
$k$ & Signed integer code stored for one weight, with
$k\in\{-M,\ldots,M\}$. \\
$M$ & Largest positive magnitude code, $M=2^{B-1}-1$. \\
$t_k$ & Normalized code magnitude, $|k|/M$. Valid codes sample
$\{0,1/M,\ldots,1\}$. \\
$s$ & Positive group scale. Codes $k=+M$ and $k=-M$ reconstruct to $+s$ and
$-s$, respectively. \\
$a$ & Stored dimensionless shape coefficient and initial slope, $a=q'(0)$. \\
$b$ & Stored dimensionless shape coefficient and half the initial curvature,
$b=q''(0)/2$. \\
$c$ & Derived cubic coefficient, $1-a-b$; it is not stored. \\
$q(t)$ & Continuous normalized-magnitude map from $[0,1]$ to $[0,1]$. Actual
reconstruction uses only the discrete coordinates $t_k$. \\
$\widehat{w}(k)$ & Signed reconstructed value represented by code $k$. \\
\bottomrule
\end{tabularx}
\endgroup

The factor \(t\) gives \(q(0)=0\), while \(c=1-a-b\) gives \(q(1)=1\).
Thus zero and both group endpoints are exact by construction. The two
stored shape coefficients jointly redistribute the interior
reconstruction levels. Candidate shape parameters must also satisfy

\begin{equation}
q'(t)=a+2bt+3(1-a-b)t^2 > 0,
\qquad t\in[0,1].
\label{eq:monotonicity}
\end{equation}

which prevents level inversions. Strict monotonicity can be checked
cheaply because the derivative is quadratic.

Uniform symmetric integer quantization is not merely a baseline adjacent
to the format. It is a point inside the family:

\begin{equation}
a=1,\qquad b=0,\qquad c=0
\quad\Longrightarrow\quad q(t)=t.
\label{eq:int-special-case}
\end{equation}

Including this point in every fitting search ensures that, before
metadata rounding, a Cubic candidate need not be worse than the
corresponding optimally clipped uniform-integer candidate under the same
objective.

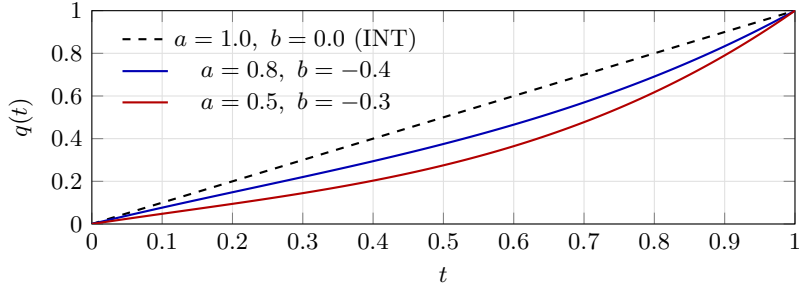
\begin{figure}[htbp]
\centering
\begin{tikzpicture}
\begin{axis}[
  width=0.68\linewidth,
  height=4.4cm,
  xmin=0, xmax=1,
  ymin=0, ymax=1,
  xlabel={$t$},
  ylabel={$q(t)$},
  grid=major,
  grid style={draw=black!12},
  legend style={font=\small,draw=none,fill=none,at={(0.03,0.97)},anchor=north west},
  tick label style={font=\small},
  label style={font=\small}
]
\addplot[black,thick,dashed,domain=0:1,samples=101] {x};
\addlegendentry{$a=1.0,\ b=0.0$ (INT)}
\addplot[blue!70!black,thick,domain=0:1,samples=101] {0.8*x-0.4*x^2+0.6*x^3};
\addlegendentry{$a=0.8,\ b=-0.4$}
\addplot[red!70!black,thick,domain=0:1,samples=101] {0.5*x-0.3*x^2+0.8*x^3};
\addlegendentry{$a=0.5,\ b=-0.3$}
\end{axis}
\end{tikzpicture}
\caption{Illustrative members of the normalized Cubic level family. All
curves satisfy $q(0)=0$ and $q(1)=1$; the shape parameters alter only the
interior level placement.}
\label{fig:cubic-shape-family}
\end{figure}

Figure \ref{fig:cubic-shape-family} illustrates how the two shape
coefficients redistribute levels without changing either endpoint.
Curves below the linear grid assign smaller reconstructed magnitudes to
a given interior code index, thereby concentrating more discrete levels
near zero while preserving the group scale at \(t=1\).

\hypertarget{metadata-and-effective-width}{%
\subsection{Metadata and effective
width}\label{metadata-and-effective-width}}

Let a Linear weight matrix have logical shape \(N\times K\). CubicQuant
uses a \([1,G]\) group shape: each output row is divided independently
into contiguous groups of \(G\) input-channel weights. Group boundaries
are identical across rows, but the fitted metadata is not shared. The
three metadata tensors therefore have logical shape \(N\times(K/G)\)
when \(K\) is divisible by \(G\).

Each group stores \(GB\) payload bits, one FP32 scale \(s\), and two
FP16 shape coefficients \(a\) and \(b\).

The metadata is therefore eight bytes per group. Ignoring row-tail
padding and container metadata, its contribution is

\begin{equation}
\frac{N(K/G)\,64}{NK}=\frac{64}{G}
\qquad\text{bits per weight},
\end{equation}

and the effective stored width is

\begin{equation}
B_{\mathrm{eff}} = B + \frac{64}{G}
\qquad\text{bits per weight}.
\label{eq:effective-bits}
\end{equation}

The following comparison places this layout beside two representative
weight scaling schemes. It describes both metadata geometry and
effective stored width, not a universal rule for all integer or
floating-point formats.

\begingroup
\renewcommand{\arraystretch}{1.12}
\noindent\begin{tabularx}{\linewidth}{@{}l c c >{\raggedright\arraybackslash}X l@{}}
\toprule
\textbf{Scheme} & \textbf{Payload} & \textbf{Scale region} &
\textbf{Metadata per region} & \textbf{Effective width} $\downarrow$ \\
\midrule
Symmetric groupwise INT & $B$ & $[1,G]$ & one FP16/BF16 scale &
$B+16/G$ \\
CubicQuant & $B$ & $[1,G]$ & one FP32 scale and two FP16 shape values &
$B+64/G$ \\
2-D block-scaled FP8 & $8$ & $[G,G]$ & one FP32 scale &
$8+32/G^2$ \\
\bottomrule
\end{tabularx}
\endgroup

The \([1,G]\) integer layout is the common groupwise weight-only case;
asymmetric variants additionally store a zero point. The FP8 row
represents two-dimensional \([G,G]\) weight scaling; activation scaling
may instead use \([1,G]\) tiles. Its smaller scale overhead is not a
like-for-like representation comparison: each FP8 payload already
carries exponent and mantissa fields, whereas INT and Cubic payloads
rely more directly on group-local metadata.

At the same value of \(G\), the effective widths are:

\begin{longtable}[]{@{}rrrr@{}}
\toprule
\begin{minipage}[b]{0.22\columnwidth}\raggedleft
\(G\)\strut
\end{minipage} & \begin{minipage}[b]{0.22\columnwidth}\raggedleft
Symmetric INT, \([1,G]\) \(\downarrow\)\strut
\end{minipage} & \begin{minipage}[b]{0.22\columnwidth}\raggedleft
CubicQuant, \([1,G]\) \(\downarrow\)\strut
\end{minipage} & \begin{minipage}[b]{0.22\columnwidth}\raggedleft
FP8, \([G,G]\) \(\downarrow\)\strut
\end{minipage}\tabularnewline
\midrule
\endhead
\begin{minipage}[t]{0.22\columnwidth}\raggedleft
128\strut
\end{minipage} & \begin{minipage}[t]{0.22\columnwidth}\raggedleft
\(B+0.12500\)\strut
\end{minipage} & \begin{minipage}[t]{0.22\columnwidth}\raggedleft
\(B+0.500\)\strut
\end{minipage} & \begin{minipage}[t]{0.22\columnwidth}\raggedleft
\(8+0.001953\)\strut
\end{minipage}\tabularnewline
\begin{minipage}[t]{0.22\columnwidth}\raggedleft
256\strut
\end{minipage} & \begin{minipage}[t]{0.22\columnwidth}\raggedleft
\(B+0.06250\)\strut
\end{minipage} & \begin{minipage}[t]{0.22\columnwidth}\raggedleft
\(B+0.250\)\strut
\end{minipage} & \begin{minipage}[t]{0.22\columnwidth}\raggedleft
\(8+0.000488\)\strut
\end{minipage}\tabularnewline
\begin{minipage}[t]{0.22\columnwidth}\raggedleft
512\strut
\end{minipage} & \begin{minipage}[t]{0.22\columnwidth}\raggedleft
\(B+0.03125\)\strut
\end{minipage} & \begin{minipage}[t]{0.22\columnwidth}\raggedleft
\(B+0.125\)\strut
\end{minipage} & \begin{minipage}[t]{0.22\columnwidth}\raggedleft
\(8+0.000122\)\strut
\end{minipage}\tabularnewline
\bottomrule
\end{longtable}

Larger groups amortize metadata and codebook-generation work, but
provide one curve for a wider region of the tensor. Whether the
additional shape freedom is enough to compensate for that coarser
granularity is an empirical question; it is tested rather than assumed.

The shape parameters use FP16, whereas the scale remains FP32. This
asymmetry keeps the dominant magnitude in higher precision while
allowing the two dimensionless shape coefficients to contribute only
four metadata bytes per group.

\hypertarget{quantization-distortion-under-reference-distributions}{%
\section{Quantization distortion under reference
distributions}\label{quantization-distortion-under-reference-distributions}}

This section gives a population reference for the finite-group
experiment in Section 3.6. Let \(X\) be a zero-mean, unit-variance
symmetric random variable with density \(f(x)\), and let \(Q\) be a
scalar quantizer. Its mean-squared distortion is

\begin{equation}
D(Q)=\mathbb{E}\!\left[(X-Q(X))^2\right].
\label{eq:population-risk}
\end{equation}

Throughout this section, \(Q\) reconstructs the continuous Cubic levels
\(sq(t_k)\). Accordingly, \(D_{\rm C}\) and every numerical value
reported below are pre-carrier population references. They characterize
the representation itself before the additional INT8 grid projection
used by Dynamic-A8 execution. Section 4.3 defines the joint
continuous/carrier objective adopted for dual-path representation
fitting; its numerical optimum is a different quantity and must not be
inferred from the tables in this section.

We first derive an exact expression in which the Cubic shape parameters
are explicit. We then bound the best Cubic quantizer between a free
Lloyd--Max codebook and the uniform-integer member of the Cubic family.

\hypertarget{exact-distortion-and-explicit-shape-dependence}{%
\subsection{Exact distortion and explicit shape
dependence}\label{exact-distortion-and-explicit-shape-dependence}}

For \(B>1\), write \(M=2^{B-1}-1\), \(t_i=i/M\), and define three fixed
basis terms

\begin{equation}
r_i=t_i^3,\qquad u_i=t_i-t_i^3,\qquad v_i=t_i^2-t_i^3.
\label{eq:cubic-basis}
\end{equation}

Substituting \(c=1-a-b\) into the level function and collecting the
coefficients of \(a\) and \(b\) gives

\begin{equation}
q(t)=at+bt^2+(1-a-b)t^3
=t^3+a(t-t^3)+b(t^2-t^3).
\label{eq:cubic-affine-expansion}
\end{equation}

Evaluating this identity at \(t_i\) makes each stored level affine in
the two shape parameters:

\begin{equation}
y_i=s\,q(t_i)=s\left(r_i+a u_i+b v_i\right).
\label{eq:affine-cubic-level}
\end{equation}

The nearest-neighbor boundary between adjacent levels is therefore

\begin{equation}
h_i=\frac{s}{2}\left[
r_i+r_{i+1}+a(u_i+u_{i+1})+b(v_i+v_{i+1})
\right].
\label{eq:affine-cubic-boundary}
\end{equation}

Equations \eqref{eq:affine-cubic-level} and
\eqref{eq:affine-cubic-boundary} show precisely where \(a\) and \(b\)
enter the distortion: they move both the reconstruction values and their
Voronoi boundaries. With \(h_{-1}=0\) and \(h_M=\infty\), symmetry gives

\begin{equation}
D(s,a,b)=2\sum_{i=0}^{M}
\int_{h_{i-1}}^{h_i}(x-y_i)^2 f(x)\,\mathrm{d}x.
\label{eq:cubic-population-risk}
\end{equation}

For each quantization cell, define its order-\(r\) cell moment

\begin{equation}
\mu_r(\ell,u)=\int_{\ell}^{u}x^r f(x)\,\mathrm{d}x,
\qquad r\in\{0,1,2\}.
\end{equation}

Expanding the square turns each cell integral into

\begin{equation}
\mu_2(h_{i-1},h_i)-2y_i\mu_1(h_{i-1},h_i)
+y_i^2\mu_0(h_{i-1},h_i).
\label{eq:cell-moment-risk}
\end{equation}

For compactness, put \(R_i(a,b)=r_i+a u_i+b v_i\). Substituting
\(y_i=sR_i\) and \(h_i=s(R_i+R_{i+1})/2\) into the preceding equation
gives the complete three-parameter objective

\begin{equation}
\boxed{
D_B(s,a,b)=2\sum_{i=0}^{M}\left[
\mu_2-2sR_i(a,b)\mu_1+s^2R_i(a,b)^2\mu_0
\right]_{(h_{i-1},h_i)} }
\label{eq:explicit-cubic-risk}
\end{equation}

where \([\mu_r]_{(h_{i-1},h_i)}\) means that every cell moment is
evaluated on that cell's two boundaries. Thus \(a\) and \(b\) are
present both explicitly in \(R_i\) and implicitly through the limits
\(h_{i-1},h_i\); treating the boundaries as fixed while changing the
curve would give the wrong distortion.

For the unit-variance reference distributions, the required
positive-half moments are closed form. Let \(A=\sqrt{3}\),
\(\beta=1/\sqrt{2}\), \(E_x=\exp(-x/\beta)\), and let \(\phi\) and
\(\Phi\) denote the standard-normal PDF and CDF. For Uniform, replace
\([\ell,u]\) by its intersection with the support \([0,A]\). Then

\begin{align}
\text{Uniform:}\quad
\mu_r(\ell,u)
&=\frac{u^{r+1}-\ell^{r+1}}{2A(r+1)}, \\
\text{Gaussian:}\quad
\mu_0&=\Phi(u)-\Phi(\ell), \\
\mu_1&=\phi(\ell)-\phi(u), \\
\mu_2&=\mu_0+\ell\phi(\ell)-u\phi(u), \\
\text{Laplace:}\quad
\mu_0&=\tfrac12(E_\ell-E_u), \\
\mu_1&=\tfrac12[(\ell+\beta)E_\ell-(u+\beta)E_u], \\
\mu_2&=\tfrac12[(\ell^2+2\beta\ell+2\beta^2)E_\ell
 -(u^2+2\beta u+2\beta^2)E_u],
\label{eq:reference-cell-moments}
\end{align}

For the unbounded final cell, the upper-endpoint terms are evaluated in
the limit \(u\to\infty\) and vanish. Substitution into
\eqref{eq:cell-moment-risk} then gives an exact evaluation of
\(D(s,a,b)\).

These expressions follow directly from elementary antiderivatives.
Uniform uses \(\int x^r\,\mathrm dx=x^{r+1}/(r+1)\). For Gaussian,
\(\phi'(x)=-x\phi(x)\) gives

\begin{equation}
\int x\phi(x)\,\mathrm dx=-\phi(x),\qquad
\int x^2\phi(x)\,\mathrm dx=\Phi(x)-x\phi(x).
\end{equation}

For the positive half of the unit-variance Laplace density,
\(f(x)=E_x/(2\beta)\); integrating \(x^rE_x\) for \(r=0,1,2\) produces
the three endpoint differences in \eqref{eq:reference-cell-moments}. No
distribution tail is discarded: the last cell is evaluated with upper
limit \(u\to\infty\).

The monotonicity condition also gives an explicit feasible domain for
the shape search. Put \(c=1-a-b\). Since \(q'(t)=a+2bt+3ct^2\), its
minimum over \([0,1]\) is attained at an endpoint or, when it lies
inside the interval, at \(t_*=-b/(3c)\). The endpoint values are
\(q'(0)=a\) and \(q'(1)=3-2a-b\). If \(c\le0\), the quadratic is concave
and its minimum on a closed interval is an endpoint. If \(c>0\), an
interior minimum exists exactly when \(0<t_*<1\), equivalently
\(-3c<b<0\), and substitution gives \(q'(t_*)=a-b^2/(3c)\). Hence

\begin{equation}
m(a,b)=
\begin{cases}
\min\!\left\{a,\,3-2a-b,\,a-\dfrac{b^2}{3c}\right\},
& c>0\ \text{and}\ -3c<b<0,\\[5pt]
\min\!\left\{a,\,3-2a-b\right\}, & \text{otherwise},
\end{cases}
\label{eq:shape-domain}
\end{equation}

The admissible closure is \(\mathcal A=\{(a,b):m(a,b)\ge0\}\). Together,
Equations \eqref{eq:affine-cubic-level}, \eqref{eq:cell-moment-risk},
and \eqref{eq:shape-domain} reduce population fitting to a constrained
optimization over only three variables: the scale \(s\) and the two
stored shape parameters \((a,b)\). The next subsection compares the
optimum of this constrained family with free and uniform codebooks.

\hypertarget{cubic-lower-and-upper-bounds}{%
\subsection{Cubic lower and upper
bounds}\label{cubic-lower-and-upper-bounds}}

Let \(\mathcal Y_B\) contain every ordered, symmetric scalar codebook
with \(2^B-1\) reconstruction levels and an exact zero. Let
\(\mathcal C_B\) contain only the codebooks generated by
\eqref{eq:affine-cubic-level} with \(s>0\) and \((a,b)\in\mathcal A\),
and let \(\mathcal I_B\) be the scaled uniform integer codebooks
obtained at \((a,b)=(1,0)\). These are nested:

\begin{equation}
\mathcal I_B\subseteq\mathcal C_B\subseteq\mathcal Y_B.
\label{eq:codebook-inclusions}
\end{equation}

The first inclusion follows by direct substitution: at \((a,b)=(1,0)\),
Equation \eqref{eq:cubic-affine-expansion} reduces to \(q(t)=t\), so
every scaled uniform-integer level set \(\{0,s/M,2s/M,\ldots,s\}\) is a
member of \(\mathcal C_B\). The second inclusion follows from
\(m(a,b)\ge0\): the Cubic levels are ordered, symmetric, contain zero,
and number \(2^B-1\), which are precisely the defining requirements of
\(\mathcal Y_B\); \(\mathcal Y_B\) simply removes the cubic-curve
restriction.

For any objective \(D\) and feasible sets
\(\mathcal S_1\subseteq\mathcal S_2\),

\begin{equation}
\inf_{z\in\mathcal S_2}D(z)\le
\inf_{z\in\mathcal S_1}D(z),
\label{eq:min-set-inclusion}
\end{equation}

because the minimization on the left may choose every candidate
available on the right and possibly more. Applying
\eqref{eq:min-set-inclusion} first to
\(\mathcal C_B\subseteq\mathcal Y_B\) and then to
\(\mathcal I_B\subseteq\mathcal C_B\) proves

\begin{equation}
\underbrace{D_{\mathrm{LM}}(B)}_{\displaystyle
\min_{Y\in\mathcal Y_B}D(Y)}
\;\le\;
\underbrace{D_{\mathrm C}(B)}_{\displaystyle
\min_{s>0,(a,b)\in\mathcal A}D(s,a,b)}
\;\le\;
\underbrace{D_{\mathrm{INT}}(B)}_{\displaystyle
\min_{s>0}D(s,1,0)}.
\label{eq:cubic-sandwich}
\end{equation}

A finite floating-point family does not extend this chain. For an
\(\mathrm{E}e\mathrm{M}m\) format with one sign bit, \(e+m=B-1\), let
all exponent/mantissa patterns denote finite values. With exponent field
\(E\), mantissa field \(J\), and bias \(\gamma_e=2^{e-1}-1\), its
nonnegative raw levels are

\begin{equation}
g_{e,m}(E,J)=
\begin{cases}
J\,2^{-m}\,2^{1-\gamma_e}, & E=0,\\
(1+J\,2^{-m})\,2^{E-\gamma_e}, & E>0.
\end{cases}
\label{eq:finite-fp-levels}
\end{equation}

Sort the distinct nonnegative values and normalize by the largest one:
\(0=z_0<z_1<\cdots<z_M=1\), where \(M=2^{B-1}-1\). With an external
scale \(s\), the positive reconstruction levels and their
nearest-neighbor boundaries are

\begin{equation}
y_i^{\mathrm{FP}}=s z_i,\qquad
\kappa_i(s)=\frac{s}{2}(z_i+z_{i+1}),\qquad
\kappa_{-1}=0,\quad\kappa_M=\infty.
\label{eq:finite-fp-boundaries}
\end{equation}

Substituting these quantities into the same cell-moment expansion used
for Cubic gives the complete FP objective

\begin{equation}
D_{e,m}(s)=2\sum_{i=0}^{M}\left[
\mu_2(\kappa_{i-1},\kappa_i)
-2s z_i\mu_1(\kappa_{i-1},\kappa_i)
+s^2z_i^2\mu_0(\kappa_{i-1},\kappa_i)
\right].
\label{eq:finite-fp-distortion}
\end{equation}

The best finite-FP reference at width \(B\) is therefore

\begin{equation}
D_{\mathrm{FP}}(B)=
\min_{e+m=B-1}\;\min_{s>0}D_{e,m}(s).
\label{eq:fp-reference}
\end{equation}

It obeys the separate bound \(D_{\mathrm{LM}}(B)\le D_{\mathrm{FP}}(B)\)
because every finite-FP codebook is also a member of the unrestricted
set \(\mathcal Y_B\). The finite-FP and Cubic families do not contain
one another, so there is no distribution-free ordering between them;
their comparison must be evaluated directly.

It remains to characterize the free-codebook lower bound. For fixed
adjacent levels \(y_i<y_{i+1}\), nearest-neighbor assignment places
their boundary where their squared errors are equal:

\begin{equation}
(h_i-y_i)^2=(h_i-y_{i+1})^2
\quad\Longrightarrow\quad
h_i=\frac{y_i+y_{i+1}}{2}.
\label{eq:lloyd-midpoint-derivation}
\end{equation}

For fixed boundaries, differentiating the distortion of cell \(i\) with
respect to its reconstruction value gives

\begin{equation}
\frac{\partial}{\partial y_i}
\int_{h_{i-1}}^{h_i}(x-y_i)^2f(x)\,\mathrm dx
=2\left[y_i\mu_0(h_{i-1},h_i)-\mu_1(h_{i-1},h_i)\right].
\label{eq:lloyd-centroid-derivation}
\end{equation}

Setting this derivative to zero yields the conditional mean. Alternating
the two necessary conditions gives the Lloyd--Max fixed point
\cite{lloyd1982,max1960}:

\begin{align}
h_i &= \frac{y_i+y_{i+1}}{2}, \\
y_i &= \mathbb{E}[X\mid h_{i-1}\le X<h_i].
\label{eq:lloyd-max}
\end{align}

The resulting codebook belongs to the largest feasible set
\(\mathcal Y_B\); its distortion is therefore the lower term in
\eqref{eq:cubic-sandwich}.

Thus \(D_{\mathrm C}-D_{\mathrm{LM}}\) measures the restriction imposed
by the two-parameter cubic curve, while
\(D_{\mathrm{INT}}-D_{\mathrm C}\) measures the benefit of those two
parameters over a linear grid. We report the fraction of the available
gap closed by Cubic as

\begin{equation}
\eta_B=\frac{D_{\mathrm{INT}}-D_{\mathrm C}}
{D_{\mathrm{INT}}-D_{\mathrm{LM}}}.
\label{eq:gap-closed}
\end{equation}

Equation \eqref{eq:cubic-sandwich} ensures \(0\le\eta_B\le1\) whenever
the denominator is nonzero. This is a bound on the \emph{optimized Cubic
family}, not a claim that every arbitrary \((a,b)\) choice is better
than INT.

\hypertarget{population-solutions}{%
\subsection{Population solutions}\label{population-solutions}}

For \(X\sim\mathrm{Uniform}[-\sqrt3,\sqrt3]\), the optimal odd
\(L=2^B-1\) level codebook has equal-width cells of width
\(\Delta=2\sqrt3/L\). The outer positive reconstruction point is the
center of the last cell, \(s^*=\sqrt3-\Delta/2=\sqrt3(1-1/L)\). Because
the density is \(1/(2\sqrt3)\), translating every cell to
\([-\Delta/2,\Delta/2]\) gives

\begin{equation}
D^*=\frac{L}{2\sqrt3}
\int_{-\Delta/2}^{\Delta/2}x^2\,\mathrm dx
=\frac{L\Delta^3}{24\sqrt3}=\frac1{L^2}.
\label{eq:uniform-distortion-derivation}
\end{equation}

The cell centers are uniformly spaced, so the corresponding Cubic
parameters are \(a^*=1,b^*=0\). Collecting the result,

\begin{equation}
s^*=\sqrt3\left(1-\frac1L\right),\qquad
a^*=1,\quad b^*=0,\qquad
D^*=\frac1{L^2}.
\label{eq:uniform-solution}
\end{equation}

Therefore all three terms in \eqref{eq:cubic-sandwich} are equal for
Uniform at W2--W8. Cubic cannot improve the population optimum there;
any finite-group improvement comes from adapting to sample fluctuations
within individual groups.

W3 provides a second result that follows from the parameter count rather
than from a numerical coincidence. Here \(M=3\) and a free symmetric
codebook has three positive levels \(0<y_1<y_2<y_3\). Set
\(p_1=y_1/y_3\), \(p_2=y_2/y_3\), and \(s=y_3\). Evaluating
\eqref{eq:cubic-affine-expansion} at \(t=1/3\) and \(t=2/3\) gives

\begin{equation}
27p_1=1+8a+2b,\qquad
27p_2=8+10a+4b.
\end{equation}

Solving this two-by-two system yields

\begin{equation}
a=1+9p_1-\frac92p_2,\qquad
b=18p_2-\frac{45}{2}p_1-\frac92.
\label{eq:w3-shape-solution}
\end{equation}

Thus every W3 free-codebook solution whose recovered \((a,b)\) lies in
\(\mathcal A\) is represented exactly by Cubic. The Gaussian and Laplace
Lloyd--Max solutions below satisfy this feasibility check, so equality
at W3 is explained by \eqref{eq:w3-shape-solution}; it is not inferred
from rounded table entries.

For Gaussian and Laplace, we evaluated the cell moments in
\eqref{eq:reference-cell-moments}, solved Lloyd--Max to convergence,
minimized the Cubic objective over \(s\) and the feasible \((a,b)\)
domain, independently optimized the INT scale, and enumerated all
finite-FP splits in \eqref{eq:fp-reference}. All distributions have unit
variance; values are MSE. Concretely, every candidate split uses levels
from \eqref{eq:finite-fp-levels}, evaluates
\eqref{eq:finite-fp-distortion}, minimizes its single scale variable,
and competes in the outer minimum of \eqref{eq:fp-reference}. The best
finite-FP reference table reports that outer minimum and its winning
split.

The finite-FP comparison uses a different denominator from
\eqref{eq:gap-closed}. Its final column is the direct relative MSE
reduction

\begin{equation}
\rho_B^{\mathrm{FP}}
=\frac{D_{\mathrm{FP}}-D_{\mathrm C}}{D_{\mathrm{FP}}}.
\label{eq:fp-reduction}
\end{equation}

\vspace{0.65\baselineskip}
\begingroup
\small
\begin{center}
\begin{minipage}{0.92\linewidth}
\centering
\textbf{Gaussian}\\[3pt]
\begin{tabular*}{\linewidth}{@{\extracolsep{\fill}}crrrrrr@{}}
\toprule
$B$ & $a^*$ & $b^*$ & $D_{\rm LM}\downarrow$ & $D_{\rm C}\downarrow$ &
$D_{\rm INT}\downarrow$ & \shortstack{\textbf{INT--LM gap}\\\textbf{recovered} $\uparrow$} \\
\midrule
2 & 1.0000 &  0.0000 & 0.190174 & 0.190174 & 0.190174 & -- \\
3 & 0.8517 & -0.1852 & 0.0440004 & 0.0440004 & 0.0468600 & 100.0\% \\
4 & 0.7932 & -0.3847 & 0.0107372 & 0.0108210 & 0.0128894 & 96.1\% \\
5 & 0.7510 & -0.5253 & 0.00266411 & 0.00275245 & 0.00369322 & 91.4\% \\
6 & 0.7192 & -0.6213 & 0.000664545 & 0.000712873 & 0.00106929 & 88.1\% \\
7 & 0.6933 & -0.6877 & 0.000166043 & 0.000186029 & 0.000308622 & 86.0\% \\
8 & 0.6711 & -0.7346 & 0.0000415075 & 0.0000486179 & 0.0000883080 & 84.8\% \\
\bottomrule
\end{tabular*}
\end{minipage}
\end{center}

\begin{center}
\begin{minipage}{0.92\linewidth}
\centering
\textbf{Laplace}\\[3pt]
\begin{tabular*}{\linewidth}{@{\extracolsep{\fill}}crrrrrr@{}}
\toprule
$B$ & $a^*$ & $b^*$ & $D_{\rm LM}\downarrow$ & $D_{\rm C}\downarrow$ &
$D_{\rm INT}\downarrow$ & \shortstack{\textbf{INT--LM gap}\\\textbf{recovered} $\uparrow$} \\
\midrule
2 & 1.0000 &  0.0000 & 0.264241 & 0.264241 & 0.264241 & -- \\
3 & 0.6216 & -0.1495 & 0.0680875 & 0.0680875 & 0.0830865 & 100.0\% \\
4 & 0.5203 & -0.2936 & 0.0172934 & 0.0176090 & 0.0273952 & 96.9\% \\
5 & 0.4603 & -0.3881 & 0.00435851 & 0.00464288 & 0.00908513 & 94.0\% \\
6 & 0.4199 & -0.4519 & 0.00109410 & 0.00123904 & 0.00297951 & 92.3\% \\
7 & 0.3892 & -0.4938 & 0.000274090 & 0.000331995 & 0.000959943 & 91.6\% \\
8 & 0.3642 & -0.5208 & 0.0000685935 & 0.0000888644 & 0.000303302 & 91.4\% \\
\bottomrule
\end{tabular*}
\end{minipage}
\end{center}

\begin{center}
\begin{minipage}{0.92\linewidth}
\centering
\textbf{Best finite-FP reference}\\[3pt]
\begin{tabular*}{\linewidth}{@{\extracolsep{\fill}}lcrrrr@{}}
\toprule
\textbf{Distribution} & $B$ & \textbf{Split} & $D_{\rm FP}\downarrow$ &
$D_{\rm C}\downarrow$ &
\shortstack{\textbf{MSE reduction}\\\textbf{vs. FP} $\uparrow$} \\
\midrule
Gaussian & 2 & E1M0 & 0.190174 & 0.190174 & 0.00\% \\
         & 3 & E1M1 & 0.0468600 & 0.0440004 & 6.10\% \\
         & 4 & E2M1 & 0.0126849 & 0.0108210 & 14.69\% \\
         & 5 & E2M2 & 0.00325600 & 0.00275245 & 15.47\% \\
         & 6 & E2M3 & 0.000829147 & 0.000712873 & 14.02\% \\
         & 7 & E2M4 & 0.000211523 & 0.000186029 & 12.05\% \\
         & 8 & E2M5 & 0.0000541207 & 0.0000486179 & 10.17\% \\
\addlinespace[2pt]
Laplace  & 2 & E1M0 & 0.264241 & 0.264241 & 0.00\% \\
         & 3 & E2M0 & 0.0706815 & 0.0680875 & 3.67\% \\
         & 4 & E2M1 & 0.0184910 & 0.0176090 & 4.77\% \\
         & 5 & E2M2 & 0.00501991 & 0.00464288 & 7.51\% \\
         & 6 & E2M3 & 0.00139751 & 0.00123904 & 11.34\% \\
         & 7 & E2M4 & 0.000396214 & 0.000331995 & 16.21\% \\
         & 8 & E2M5 & 0.000113607 & 0.0000888644 & 21.78\% \\
\bottomrule
\end{tabular*}
\end{minipage}
\end{center}
\par\medskip
\endgroup

At W2 there is no interior positive level, so \((a,b)\) is not
identifiable; the table reports the canonical INT point. At W3,
\(s,a,b\) provide exactly three degrees of freedom for the three
positive levels, and the optimized Cubic codebook reaches the Lloyd--Max
lower bound for both tested unbounded distributions. From W4 onward,
\(D_{\mathrm C}>D_{\mathrm{LM}}\) exposes the cost of describing more
than three positive levels with only \(s,a,b\). Even so, Cubic closes
84.8--96.1\% of the Gaussian gap and 91.4--96.9\% of the Laplace gap
over W4--W8. The decreasing \(a^*\) and increasingly negative \(b^*\)
move interior levels toward the high-density center while preserving a
more distant endpoint for the tails.

For Uniform, the finite-FP split E1M\((B-2)\) is the same linear grid as
INT and therefore also attains \(1/(2^B-1)^2\). For Gaussian and
Laplace, Cubic is better than the best enumerated finite-FP split at
every width above W2 in this population comparison, but this is a
numerical result for the stated distributions rather than a consequence
of set inclusion.

\hypertarget{numerical-verification-of-analytic-distortion}{%
\subsection{Numerical verification of analytic
distortion}\label{numerical-verification-of-analytic-distortion}}

An independent Monte Carlo estimate checks the analytic distortion in
the preceding subsection. For each distribution, 2,000,000 samples are
assigned to their nearest reconstruction levels and their empirical
squared error is compared with the closed-form cell-moment result.
Across all 21 distribution--bit-width combinations, the largest
discrepancy is 2.10 estimated standard errors. Here \(D_{\rm analytic}\)
is the optimized Cubic population value \(D_{\rm C}\) and SE is the
estimated standard error of the Monte Carlo mean.

\begingroup
\small
\setlength{\tabcolsep}{5.8pt}
\begin{center}
\begin{minipage}{0.96\linewidth}
\centering
\textbf{Uniform}\\[3pt]
\begin{tabular*}{\linewidth}{@{\extracolsep{\fill}}crrrrr@{}}
\toprule
$B$ & $D_{\rm analytic}\downarrow$ & $D_{\rm MC}\downarrow$ & \textbf{MC SE} $\downarrow$ &
$(D_{\rm analytic}-D_{\rm MC})/\mathrm{SE}\to0$ & \textbf{Relative difference} $\downarrow$ \\
\midrule
2 & 0.111111111 & 0.111176809 & $7.03\!\times\!10^{-5}$ & -0.94 & 0.059\% \\
3 & 0.0204081633 & 0.0204092026 & $1.29\!\times\!10^{-5}$ & -0.08 & 0.005\% \\
4 & 0.00444444444 & 0.00444717327 & $2.81\!\times\!10^{-6}$ & -0.97 & 0.061\% \\
5 & 0.00104058273 & 0.00104130287 & $6.58\!\times\!10^{-7}$ & -1.09 & 0.069\% \\
6 & 0.000251952633 & 0.000252070370 & $1.59\!\times\!10^{-7}$ & -0.74 & 0.047\% \\
7 & 0.000062000124 & 0.000062005266 & $3.92\!\times\!10^{-8}$ & -0.13 & 0.008\% \\
8 & 0.000015378701 & 0.000015369324 & $9.72\!\times\!10^{-9}$ & +0.96 & 0.061\% \\
\bottomrule
\end{tabular*}
\end{minipage}
\end{center}

\begin{center}
\begin{minipage}{0.96\linewidth}
\centering
\textbf{Gaussian}\\[3pt]
\begin{tabular*}{\linewidth}{@{\extracolsep{\fill}}crrrrr@{}}
\toprule
$B$ & $D_{\rm analytic}\downarrow$ & $D_{\rm MC}\downarrow$ & \textbf{MC SE} $\downarrow$ &
$(D_{\rm analytic}-D_{\rm MC})/\mathrm{SE}\to0$ & \textbf{Relative difference} $\downarrow$ \\
\midrule
2 & 0.190174039 & 0.190646342 & $2.58\!\times\!10^{-4}$ & -1.83 & 0.248\% \\
3 & 0.0440003825 & 0.0441182339 & $7.85\!\times\!10^{-5}$ & -1.50 & 0.268\% \\
4 & 0.0108210079 & 0.0108585225 & $2.62\!\times\!10^{-5}$ & -1.43 & 0.347\% \\
5 & 0.00275245089 & 0.00276697900 & $9.92\!\times\!10^{-6}$ & -1.46 & 0.528\% \\
6 & 0.000712872730 & 0.000719439447 & $4.27\!\times\!10^{-6}$ & -1.54 & 0.921\% \\
7 & 0.000186029096 & 0.000189446766 & $1.96\!\times\!10^{-6}$ & -1.75 & 1.837\% \\
8 & 0.000048617883 & 0.000049916906 & $8.87\!\times\!10^{-7}$ & -1.47 & 2.672\% \\
\bottomrule
\end{tabular*}
\end{minipage}
\end{center}

\begin{center}
\begin{minipage}{0.96\linewidth}
\centering
\textbf{Laplace}\\[3pt]
\begin{tabular*}{\linewidth}{@{\extracolsep{\fill}}crrrrr@{}}
\toprule
$B$ & $D_{\rm analytic}\downarrow$ & $D_{\rm MC}\downarrow$ & \textbf{MC SE} $\downarrow$ &
$(D_{\rm analytic}-D_{\rm MC})/\mathrm{SE}\to0$ & \textbf{Relative difference} $\downarrow$ \\
\midrule
2 & 0.264241118 & 0.265203848 & $6.30\!\times\!10^{-4}$ & -1.53 & 0.364\% \\
3 & 0.0680875037 & 0.0684364227 & $2.37\!\times\!10^{-4}$ & -1.47 & 0.512\% \\
4 & 0.0176089626 & 0.0177324582 & $9.05\!\times\!10^{-5}$ & -1.36 & 0.701\% \\
5 & 0.00464287524 & 0.00467610756 & $3.59\!\times\!10^{-5}$ & -0.93 & 0.716\% \\
6 & 0.00123903759 & 0.00123766787 & $1.40\!\times\!10^{-5}$ & +0.10 & 0.111\% \\
7 & 0.000331995019 & 0.000327824299 & $4.92\!\times\!10^{-6}$ & +0.85 & 1.256\% \\
8 & 0.000088864369 & 0.000086346622 & $1.20\!\times\!10^{-6}$ & +2.10 & 2.833\% \\
\bottomrule
\end{tabular*}
\end{minipage}
\end{center}
\endgroup

The larger relative percentages occur only where W8 MSE is very small
and tail events dominate sampling variance; the standardized
discrepancies remain within 2.10 standard errors.

\hypertarget{from-population-optima-to-finite-group-fitting}{%
\subsection{From population optima to finite-group
fitting}\label{from-population-optima-to-finite-group-fitting}}

The preceding results optimize one continuous-level quantizer for an
entire reference distribution. A groupwise continuous fitter instead
selects a separate scale and shape for every finite group of weights.
These are different optimization problems, even when the weights in a
group are independent samples from the same distribution.

Let \(S_G=(X_1,\ldots,X_G)\) denote a group of \(G\) values, let
\(\theta=(s,a,b)\) collect the continuous Cubic parameters, and let
\(\ell_{\rm C}(X,\theta)\) include assignment to the nearest continuous
level generated by \(\theta\). Define the empirical distortion of one
group as

\begin{equation}
\widehat D_G^{\rm C}(\theta;S_G)
=\frac{1}{G}\sum_{j=1}^{G}\ell_{\rm C}(X_j,\theta),
\qquad
\widehat\theta(S_G)\in
\arg\min_\theta \widehat D_G^{\rm C}(\theta;S_G).
\label{eq:empirical-group-distortion}
\end{equation}

The expected distortion after independently fitting every group is
therefore

\begin{equation}
R_G^{\rm C}
=\mathbb{E}_{S_G}\!\left[
\min_\theta \widehat D_G^{\rm C}(\theta;S_G)
\right].
\label{eq:finite-group-risk}
\end{equation}

By contrast, Section 3.3 reports the population optimum

\begin{equation}
D_{{\rm pop},\rm C}^{*}
=\min_\theta \mathbb{E}_{X}[\ell_{\rm C}(X,\theta)].
\label{eq:population-optimum}
\end{equation}

If \(\theta^{*}\) is a population minimizer, then for every observed
group \(S_G\),

\begin{equation}
\min_\theta \widehat D_G^{\rm C}(\theta;S_G)
\leq \widehat D_G^{\rm C}(\theta^{*};S_G).
\end{equation}

Taking expectations on both sides gives

\begin{equation}
R_G^{\rm C}
\leq \mathbb{E}_{S_G}[\widehat D_G^{\rm C}(\theta^{*};S_G)]
=\mathbb{E}_{X}[\ell_{\rm C}(X,\theta^{*})]
=D_{{\rm pop},\rm C}^{*}.
\label{eq:finite-group-adaptation-bound}
\end{equation}

This inequality is an in-group reconstruction statement for static
weights, not a claim about out-of-sample generalization. It says that a
fitter allowed to adapt \((s,a,b)\) to each realized group cannot have
larger expected reconstruction error than a single population-wide
parameter choice. Under the usual law-of-large-numbers and regularity
conditions, the empirical objective approaches the population objective
as \(G\) grows, so this finite-group adaptation advantage diminishes.

This distinction resolves an otherwise apparent contradiction in the
Uniform result. Section 3.3 shows that the linear INT grid is population
optimal for a perfectly Uniform distribution. A finite group, however,
is not exactly Uniform: its empirical histogram, extrema, and spacing
fluctuate. Cubic can adapt its two shape parameters to those
fluctuations and may obtain a small groupwise improvement even though it
has no population-level advantage for Uniform data.

Group size consequently controls two opposing effects. Smaller groups
permit more local adaptation but require more scale and shape metadata
per weight; larger groups amortize metadata and codebook-generation cost
but force one curve to represent a broader region of the tensor. Real
weight groups are also neither independent nor stationary, so the size
of this tradeoff cannot be inferred from the reference distributions
alone. Finite-\(G\) simulation and tensor-level experiments are
therefore used to measure this finite-group effect, while the population
analysis supplies the distribution-controlled reference point. The same
empirical-versus-population argument applies when the loss is replaced
consistently by the joint carrier-aware loss in Section 4.3, but the
resulting optimum and decision cells are not the continuous quantities
tabulated in this section.

\hypertarget{finite-group-reconstruction-experiment}{%
\subsection{Finite-group reconstruction
experiment}\label{finite-group-reconstruction-experiment}}

The population results above use one parameter set for an entire
distribution. To measure the distinct finite-group setting of Section
3.5, we drew 15,360 zero-centered values from each reference
distribution with seed 42 and divided them into groups of \(G=128\).
Every group was fitted independently. Cubic used joint scale/shape
fitting with optimized clipping; symmetric INT independently optimized
its clipping scale; and the finite-FP reference enumerated every split
\(e+m=B-1\), optimized the scale of each candidate, and retained the
lowest-error split.

Because every candidate optimizes an external scale, the experiment is
invariant to a common positive rescaling of the source values. The table
therefore reports the scale-free quantity

\begin{equation}
\operatorname{NRMSE}
=\frac{\sqrt{N^{-1}\sum_{j=1}^{N}(x_j-\widehat{x}_j)^2}}
       {\sigma_X},
\label{eq:finite-group-nrmse}
\end{equation}

where \(N=15{,}360\) and each \(\widehat{x}_j\) is reconstructed with
the parameters fitted to its own group. Lower is better; boldface marks
the unique minimum in each row, while exact ties are left unbolded. The
experiment records W2--W6 and W8; the absent W1 and W7 rows are not
interpolated. As elsewhere in Section 3, these are continuous-level
reconstructions before Dynamic-A8 carrier projection.

\begingroup
\small
\begin{center}
\begin{tabular*}{0.92\linewidth}{@{\extracolsep{\fill}}lcrrrc@{}}
\toprule
\textbf{Distribution} & $B$ & \textbf{Cubic} $\downarrow$ &
\textbf{Clipped INT} $\downarrow$ & \textbf{Finite FP} $\downarrow$ &
\textbf{FP split} \\
\midrule
Uniform  & 2 & 0.331403 & 0.331403 & 0.331403 & E1M0 \\
         & 3 & \textbf{0.139198} & 0.141685 & 0.141685 & E1M1 \\
         & 4 & \textbf{0.062979} & 0.065536 & 0.065536 & E1M2 \\
         & 5 & \textbf{0.029996} & 0.031351 & 0.031351 & E1M3 \\
         & 6 & \textbf{0.014636} & 0.015359 & 0.015359 & E1M4 \\
         & 8 & \textbf{0.003563} & 0.003792 & 0.003792 & E1M6 \\
\addlinespace[3pt]
Gaussian & 2 & 0.434638 & 0.434638 & 0.434638 & E1M0 \\
         & 3 & \textbf{0.201392} & 0.213533 & 0.210783 & E2M0 \\
         & 4 & \textbf{0.092396} & 0.106803 & 0.102027 & E2M1 \\
         & 5 & \textbf{0.043095} & 0.053124 & 0.051308 & E2M2 \\
         & 6 & \textbf{0.020502} & 0.026055 & 0.025654 & E2M3 \\
         & 8 & \textbf{0.004920} & 0.006408 & 0.006358 & E2M5 \\
\addlinespace[3pt]
Laplace  & 2 & 0.513876 & 0.513876 & 0.513876 & E1M0 \\
         & 3 & \textbf{0.236439} & 0.280624 & 0.251182 & E2M0 \\
         & 4 & \textbf{0.106764} & 0.148571 & 0.113906 & E2M1 \\
         & 5 & \textbf{0.048854} & 0.074182 & 0.054661 & E2M2 \\
         & 6 & \textbf{0.023584} & 0.036513 & 0.026644 & E2M3 \\
         & 8 & \textbf{0.005614} & 0.009004 & 0.006547 & E2M5 \\
\bottomrule
\end{tabular*}
\end{center}
\endgroup

W2 has no movable interior positive level, so all three families
coincide in this experiment. From W3 upward, per-group shape adaptation
consistently reduces reconstruction error. The improvement over clipped
INT is modest for bounded Uniform samples and larger for Gaussian and
Laplace samples. At W4, the Cubic NRMSE reductions are 3.90\%, 13.49\%,
and 28.14\%, respectively. Against the best finite-FP candidate, the
corresponding reductions are 3.90\%, 9.44\%, and 6.27\%.

Unlike the population result, the small Uniform gain does not imply that
a nonlinear grid is optimal for a Uniform law. It is the finite-sample
adaptation effect derived in Section 3.5: each realized group departs
slightly from the population distribution, and Cubic can fit those local
fluctuations. The experiment therefore supplies a finite-\(G\)
counterpart to the population tables, not evidence about CUDA throughput
or model-level task quality. Since only \(G=128\) is measured here, it
also does not determine the best group size.

\hypertarget{groupwise-parameter-fitting}{%
\section{Groupwise parameter
fitting}\label{groupwise-parameter-fitting}}

\hypertarget{continuous-reference-objective}{%
\subsection{Continuous reference
objective}\label{continuous-reference-objective}}

The representation becomes a quantizer only after codes and group
parameters have been selected. The reference estimator operates
independently on each static weight group and minimizes original-domain
squared reconstruction error over codes, scale, and admissible shape
parameters:

\begin{equation}
\min_{s,a,b,\{k_j\}}
\sum_j\left[
w_j-\operatorname{sign}(k_j)sq\!\left(\frac{|k_j|}{M}\right)
\right]^2.
\label{eq:offline-objective}
\end{equation}

Here \(s\) is both the endpoint and the clipping threshold; it is not
constrained to equal the largest magnitude in the group. Values beyond
\(s\) saturate at the signed endpoint. For fixed levels, each value is
assigned to its nearest reconstruction. The reference search traverses a
deterministic set of monotonic \((a,b)\) candidates, jointly refines
scale and clipping, and always includes the uniform-integer point. The
objective is reevaluated after mapping \(s\), \(a\), and \(b\) to their
serialized precisions, so parameter rounding is part of the fitted
representation rather than a post hoc perturbation.

The fitting loss is MSE/SSE. Its aggregate, scale-free reporting form is

\begin{equation}
\operatorname{NRMSE}_{\rm C}
=\sqrt{\frac{\sum_j(w_j-\widehat{w}_j)^2}{\sum_jw_j^2}},
\label{eq:nrmse}
\end{equation}

which is scale-free across tensors and bit widths. Taking the square
root is a reporting transformation; it does not alter the candidate
selected by MSE.

This continuous objective is the finite-group counterpart of the
population analysis in Section 3 and one component of the dual-path
objective introduced in Section 4.3. It depends only on the weights
themselves and therefore serves as a controlled reference estimator.
That choice does not imply that every coordinate has equal functional
importance; Section 4.2 separates the format from richer estimators that
incorporate model sensitivity.

\hypertarget{compatibility-with-data-driven-estimation}{%
\subsection{Compatibility with data-driven
estimation}\label{compatibility-with-data-driven-estimation}}

The weight-only fitter is a reference estimator, not a restriction of
the Cubic format. Let \(E=W-\widehat W\) be the weight error of a linear
operator and \(x\) its input. Unweighted weight MSE minimizes
\(\lVert E\rVert_F^2\). If representative activations are available, the
expected output error is instead

\begin{equation}
\mathbb{E}\|Ex\|_2^2
=\operatorname{tr}(E\Sigma_xE^\top),
\qquad \Sigma_x=\mathbb{E}[xx^\top].
\label{eq:covariance-weighted}
\end{equation}

A diagonal covariance becomes a coordinate-weighted reconstruction
objective. Block-diagonal, low-rank, or full covariance models retain
progressively more cross-coordinate structure at progressively greater
fitting cost. Approximate Hessians and other sensitivity estimates can
play the same role.

Changing the estimator changes which codes and \((s,a,b)\) are chosen;
it does not change the packed bitstream, metadata schema, level
function, or runtime kernel. CubicQuant can therefore be combined with
data-driven methods such as second-order or activation-aware fitting.
The experiments in this report evaluate only the data-free reference
path and make no claim that it dominates such methods.

\hypertarget{dynamic-a8-carrier-aware-fitting}{%
\subsection{Dynamic-A8 carrier-aware
fitting}\label{dynamic-a8-carrier-aware-fitting}}

The model-dtype execution path can reconstruct the continuous Cubic
value. The Dynamic-A8 path instead maps the normalized Cubic level to a
signed INT8 carrier:

\begin{equation}
r(k)=\operatorname{round}\!\left[
127\,\operatorname{sign}(k)q\!\left(\frac{|k|}{M}\right)
\right].
\label{eq:carrier}
\end{equation}

The deployed normalized level is consequently

\begin{equation}
\widetilde q(t)=\frac{\operatorname{round}(127q(t))}{127}.
\label{eq:carrier-grid}
\end{equation}

For a weight \(w_j\) assigned to code \(k_j\), define the two
reconstructions

\begin{equation}
\widehat w_j^{\rm C}
=\operatorname{sign}(k_j)sq(t_{k_j}),
\qquad
\widehat w_j^{\rm A8}
=\frac{s}{127}r(k_j)
=\operatorname{sign}(k_j)s\widetilde q(t_{k_j}).
\label{eq:continuous-and-carrier-reconstruction}
\end{equation}

The first is consumed by the model-dtype path; the second is the
effective weight represented by the signed INT8 carrier in the
Dynamic-A8 path. Their groupwise squared errors are

\begin{equation}
\mathcal L_{\rm C}=\sum_j(w_j-\widehat w_j^{\rm C})^2,
\qquad
\mathcal L_{\rm A8}=\sum_j(w_j-\widehat w_j^{\rm A8})^2.
\label{eq:continuous-and-carrier-loss}
\end{equation}

The dual-path fitting policy studied here gives the two execution paths
equal weight:

\begin{equation}
\mathcal L_{\rm joint}
=\frac{1}{2}\mathcal L_{\rm C}
+\frac{1}{2}\mathcal L_{\rm A8}.
\label{eq:carrier-aware-objective}
\end{equation}

This objective is distinct from the continuous distortion analyzed in
Section 3. To state the corresponding population quantity, put
\(z_i=q(t_i)\) and \(\widetilde z_i=\widetilde q(t_i)\). On the positive
half-line, the joint distortion is

\begin{equation}
D_{B}^{\rm joint}(s,a,b)
=2\sum_{i=0}^{M}
\int_{h_{i-1}^{\rm joint}}^{h_i^{\rm joint}}
\frac{(x-sz_i)^2+(x-s\widetilde z_i)^2}{2}
f(x)\,\mathrm dx.
\label{eq:carrier-aware-population-objective}
\end{equation}

Equating the joint costs of adjacent codes gives their decision boundary

\begin{equation}
h_i^{\rm joint}
=\frac{s}{2}
\frac{
(z_{i+1}^2+\widetilde z_{i+1}^2)
-(z_i^2+\widetilde z_i^2)}{
(z_{i+1}+\widetilde z_{i+1})
-(z_i+\widetilde z_i)}.
\label{eq:carrier-aware-boundary}
\end{equation}

Equation \ref{eq:carrier-aware-population-objective} can be evaluated
with the same cell moments as Equation \ref{eq:cell-moment-risk}, but
its boundaries are not the midpoints in Equation
\ref{eq:affine-cubic-boundary}. Moreover, the rounding operation makes
it piecewise non-smooth in \((a,b)\). The continuous population values
and bounds in Section 3 therefore remain representation references; they
are not numerical evaluations of the joint fitting policy.

Thus the carrier correction is not an additional diagnostic applied
after a continuous-only fit: it participates in the selection of codes,
scale, and shape parameters. The implementation minimizes
\(\mathcal L_{\rm C}+\mathcal L_{\rm A8}\), omitting the common factor
\(1/2\) because it cannot change the minimizer. For reporting, it
restores the mean and emits

\begin{equation}
\operatorname{NRMSE}_{\rm joint}
=\sqrt{
\frac{\mathcal L_{\rm C}+\mathcal L_{\rm A8}}
{2\sum_j w_j^2}}
\label{eq:carrier-aware-nrmse}
\end{equation}

together with the A8-only diagnostic
\(\sqrt{\mathcal L_{\rm A8}/\sum_jw_j^2}\).

The joint objective is used throughout the deterministic candidate
search. For fixed codes and shape parameters, let \(u_j=|w_j|\) and use
\(z_j=q(t_{k_j})\), \(\widetilde z_j=\widetilde q(t_{k_j})\). The shared
scale that minimizes Equation \ref{eq:carrier-aware-objective} has the
closed form

\begin{equation}
s^*
=\frac{\sum_j u_j(z_j+\widetilde z_j)}
{\sum_j(z_j^2+\widetilde z_j^2)}.
\label{eq:carrier-aware-scale}
\end{equation}

Conversely, for fixed \((s,a,b)\), each code is assigned by minimizing

\begin{equation}
(u_j-sq(t_k))^2
+(u_j-s\widetilde q(t_k))^2
\label{eq:carrier-aware-assignment}
\end{equation}

over the admissible magnitudes. Assignment and scale updates alternate
while the estimator traverses the same monotonic shape and soft-clipping
candidates as the continuous fitter. After converting \(a\) and \(b\) to
FP16 and retaining \(s\) in FP32, the final code assignment is repeated
under the same joint objective. The evaluated object is therefore the
serialized representation rather than an unrounded optimization
intermediate.

Equal weighting is a representation-design choice, not an assumption
about the frequency with which either execution path is selected, and no
averaging occurs during inference. A model-dtype kernel reconstructs
\(\widehat w^{\rm C}\), whereas a Dynamic-A8 kernel uses the carrier
represented by \(r(k)\). For W1 and W2 the available normalized levels
already lie exactly on the INT8 carrier grid, so
\(\mathcal L_{\rm C}=\mathcal L_{\rm A8}\) and the correction has no
effect.

Optimizing only the continuous curve at W3--W8 can choose parameters
that look favorable before the second grid projection but lose that
advantage in the actual A8 kernel. The joint correction remains
data-free because it models deterministic runtime arithmetic rather than
activation statistics. It changes neither code packing nor stored
metadata and adds no online fitting work; the same checkpoint can
therefore serve both model-dtype and Dynamic-A8 execution. In
particular, the correction covers only the weight-side projection onto
the INT8 carrier grid. It does not model dynamic activation quantization
or its propagation into operator outputs; doing so requires an
activation distribution or sensitivity estimator of the kind discussed
in Section 4.2.

\hypertarget{gpu-realization}{%
\section{GPU realization}\label{gpu-realization}}

\hypertarget{fused-decode-and-compute-principle}{%
\subsection{Fused decode-and-compute
principle}\label{fused-decode-and-compute-principle}}

Cubic codes are not native Tensor Core operands. Expanding an entire
matrix to BF16 or INT8 before multiplication would write and reread a
conventional weight tensor, eliminating much of the bandwidth benefit of
the packed representation. CubicQuant therefore couples reconstruction
to consumption: each output tile loads only its packed payload and group
metadata, generates a temporary model-dtype or INT8 operand, and
immediately consumes that operand in the dot-product main loop. The
expanded tile remains local to the kernel.

Two complementary kernel families instantiate this principle. Generic
kernels cover W1--W8 for model-dtype and Dynamic-A8 activations, with
separate narrow-row GEMV and two-dimensional dense organizations. Native
CUDA kernels add packed-word extraction and DP4A-style signed INT8 dot
products for selected Dynamic-A8 regimes. These families are alternative
realizations of the same operator semantics; the numerical format does
not depend on which realization is selected.

\hypertarget{packed-extraction-and-level-reconstruction}{%
\subsection{Packed extraction and level
reconstruction}\label{packed-extraction-and-level-reconstruction}}

Let \(\beta_p\) denote packed byte \(p\). For logical coordinate \(j\),
define \(p_j=\lfloor jB/8\rfloor\) and \(\delta_j=(jB)\bmod 8\). The
unsigned payload is recovered by

\begin{equation}
u_j=\left[
(\beta_{p_j}\!\gg\!\delta_j)
\;\mathbin{|}\;
(\beta_{p_j+1}\!\ll\!(8-\delta_j))
\right]\mathbin{\&}(2^B-1),
\label{eq:gpu-packed-extraction}
\end{equation}

where the second-byte contribution is omitted when the field does not
cross a byte boundary. W1 maps its single payload bit directly to
\(\{-1,+1\}\); wider codes are sign-extended, and the reserved
most-negative pattern is mapped to zero as required by the format.
Because \(GB\) is byte aligned, a weight group does not share a partial
byte with the next group's metadata domain.

The group index is \(g=\lfloor j/G\rfloor\). A tile loads \(s_g\) and,
where needed, \((a_g,b_g)\) once for the output channels it covers.
Three reconstruction strategies are used:

\begin{enumerate}
\def\labelenumi{\arabic{enumi}.}
\tightlist
\item
  W1 and W2 admit direct algebraic mappings and need no curve table.
\item
  A nonnegative Cubic table contains \(2^{B-1}\) entries, including
  zero; its signed values are obtained by applying the decoded sign.
\item
  Direct evaluation uses Horner-form fused multiply-adds,
\end{enumerate}

\begin{equation}
q(t)=t\,\operatorname{fma}\!\left(
t,\operatorname{fma}(t,c,b),a\right).
\label{eq:gpu-horner}
\end{equation}

For some routed W3/A8 kernels, the two interior positive carrier levels
are precomputed from \((a,b)\) after loading and reused as compact INT8
metadata. For W4--W8, eligible CUDA kernels construct a group-local
carrier table in shared memory, while generic kernels may construct
levels through static evaluation or evaluate the curve directly. A
simple arithmetic comparison is \(O(2^{B-1})\) work to construct the
positive table versus \(O(G)\) curve evaluations to decode a group. This
is only a cost-model term, not a dispatch law: packed-load pattern,
table reuse, register pressure, shared-memory capacity, output shape,
and GPU architecture determine the measured crossover.

\hypertarget{model-dtype-execution}{%
\subsection{Model-dtype execution}\label{model-dtype-execution}}

The model-dtype path accepts FP16 or BF16 activations and implements the
continuous reconstruction in Equation \ref{eq:cubic}. Its dense and
narrow-row tactics have different, explicit rounding points. In a dense
Triton tile, \(s_gq(t_k)\) is formed using FP32 metadata arithmetic,
converted to the activation dtype, and supplied with the activation tile
to \texttt{tl.dot}; the dot accumulator is FP32. In the narrow-row GEMV
path, activations and reconstructed normalized weights are multiplied in
FP32 and \(s_g\) is applied to the group reduction in FP32. Both store
the requested model dtype, but they need not be bitwise identical
because the dense path rounds the temporary weight operand before the
dot.

The continuous formula is therefore the semantic reference for this
execution mode. No individual model-dtype kernel is treated as an oracle
for Dynamic A8; both modes are compared independently with the
mathematical reconstruction defined by their respective equations.

\hypertarget{dynamic-a8-execution}{%
\subsection{Dynamic-A8 execution}\label{dynamic-a8-execution}}

For an activation row \(x\), the canonical path computes

\begin{equation}
\bar x=\max\!\left(\max_j|x_j|,\epsilon\right),
\qquad d_x=\frac{\bar x}{127},
\label{eq:a8-activation-scale}
\end{equation}

and

\begin{equation}
x_{8,j}=\operatorname{clip}\!\left(
\operatorname{round}_{\rm away}(x_j/d_x),-127,127\right).
\label{eq:a8-activation-quantization}
\end{equation}

Here \(\operatorname{round}_{\rm away}\) denotes nearest-integer
rounding with halfway cases away from zero, matching the
activation-quantization semantics. This is distinct from the
round-to-nearest-even operation used to construct the weight carrier
\(r(k)\) in Section 4.3. The distinction only affects exact halfway
cases but is part of the executable contract. An all-zero row uses the
positive safeguard \(\epsilon\) and produces zero activation carriers.

For weight group \(g\), the signed integer partial and reconstructed
output are

\begin{align}
P_g &= \sum_{j\in g}x_{8,j}r_g(k_j), \\
y &= \sum_g\frac{d_xs_g}{127}P_g.
\label{eq:a8-output}
\end{align}

Integer products accumulate into INT32 partials. With the supported
\(G\le512\), a full-group worst-case magnitude is bounded by

\begin{equation}
|P_g|\le G\,127^2\le512\times127^2=8{,}258{,}048<2^{31}-1.
\label{eq:a8-int32-bound}
\end{equation}

Dense kernels may form still smaller \(K\)-tile partials. Each completed
partial is converted to FP32 before multiplication by \(d_xs_g/127\);
cross-group accumulation is also FP32, followed by the requested output
cast. The activation scale is generated at runtime and is neither stored
in the checkpoint nor estimated from a calibration corpus.

After carrier construction, W1--W8 all perform one signed INT8 product
per logical weight. This equality concerns the main dot-product order
only. Their total costs remain different because packed traffic,
cross-byte extraction, carrier-table construction, direct curve
evaluation, and achievable occupancy all depend on \(B\) and \(G\).

Dynamic A8 is an operator-local compute mode, not a persistent integer
graph. An ordinary operator returns model-dtype output, and a later
operator quantizes its own input when invoked. Selected fused expert
paths can combine activation and quantization adjacent to the down
projection, but this optimization does not change the general boundary
or the checkpoint format.

\hypertarget{workload-regimes-and-routed-execution}{%
\subsection{Workload regimes and routed
execution}\label{workload-regimes-and-routed-execution}}

For an ordinary Linear operator, \(M\) is the number of activation rows.
Small \(M\) provides too little row parallelism for a conventional
matrix tile, so the narrow path assigns parallel work across output
channels and consumes one or a few rows as GEMV. As \(M\) grows, a
two-dimensional tile can reuse each decoded weight across several
activation rows and can make effective use of native dot instructions.
Thus decode-like and prefill-like shapes have different crossovers even
at identical \((B,G,N,K)\).

For a routed operator, the relevant work is the number and distribution
of active token--matrix routes rather than the input-token count alone.
Routes can be compacted and grouped to reuse a decoded tile, processed
independently with route-major outputs, or reduced within the kernel
when the layout permits it. Dense tiles remain useful when route density
is high. These choices alter work organization but not the Cubic
reconstruction or route weights.

Routing probabilities, expert selection, and expert placement remain
outside the Cubic format. The kernels consume the route map supplied by
the serving system and support both local and expert-parallel layouts
without assigning a special meaning to expert numbering.

\hypertarget{shape--and-device-dependent-kernel-selection}{%
\subsection{Shape- and device-dependent kernel
selection}\label{shape--and-device-dependent-kernel-selection}}

Performance depends on \((B,G,M,N,K)\), activation dtype, route
topology, GPU architecture, compiler stack, and kernel organization. No
single threshold in \(M\) or \(B\) reliably determines whether a narrow,
dense, native, or generic kernel is fastest across this space.
CubicQuant therefore treats kernel choice as a shape- and
device-dependent optimization problem rather than a property of the
numerical format.

For every admissible operator signature, candidate families are first
filtered by payload width, alignment, dtype, and architectural
constraints. Numerically valid candidates are then compared on
representative row-count regimes, with choices including GEMV versus
dense tiling, block height, route grouping, route-level parallelism, and
native CUDA versus generic kernels. The selected strategy is a
deterministic function of the measured execution environment and
operator signature.

This empirical selection changes neither codes nor group parameters. It
is also distinct from quantizer fitting: synthetic operands are
sufficient because the quantity being estimated is kernel latency, not
an activation distribution or model-sensitivity statistic. A
conservative implementation remains available whenever no specialized
candidate satisfies the numerical and structural constraints.

\hypertarget{experimental-evidence}{%
\section{Experimental evidence}\label{experimental-evidence}}

The evidence is organized in increasing order of system dependence.
Section 3 establishes distribution-controlled representation error; this
section first quantifies storage and computational work, then reports
isolated kernel measurements and one negative fusion result. End-to-end
serving behavior is discussed only as an attribution boundary because it
also depends on attention, communication, scheduling, and memory
management.

\hypertarget{experimental-scope-and-portability}{%
\subsection{Experimental scope and
portability}\label{experimental-scope-and-portability}}

The implementation can be compiled natively for several GPU generations,
but binary compatibility and measured performance are distinct claims:

\begin{longtable}[]{@{}lll@{}}
\toprule
GPU generation & Native target & Latency evidence in this
report\tabularnewline
\midrule
\endhead
Ampere data center & SM80 & Not measured\tabularnewline
Ampere consumer & SM86 & Not measured\tabularnewline
Ada consumer/data center & SM89 & Not measured\tabularnewline
Hopper & SM90 and SM90a & Single-H200 kernel measurements\tabularnewline
Blackwell data center & SM100 & Not measured\tabularnewline
Blackwell consumer & SM120 & Not measured\tabularnewline
\bottomrule
\end{longtable}

The isolated measurements were collected on one NVIDIA H200 with 143,771
MiB of memory and CUDA compute capability 9.0, using the SM90a target.
The storage and work model in Section 6.2 is architecture independent;
every measured latency in this report is Hopper-specific.

Native targets were verified from the produced code objects rather than
inferred solely from compiler flags \cite{nvidia-binary-utils}.
Physical-device evidence remains limited to Hopper, so the other rows
establish portability of the implementation, not performance equivalence
across architectures.

\hypertarget{computational-work-and-memory-traffic}{%
\subsection{Computational work and memory
traffic}\label{computational-work-and-memory-traffic}}

Consider a matrix multiplication with \(M\) input rows, \(N\) output
columns, input dimension \(K\), weight width \(B\), and weight group
size \(G\). Ignoring row-tail padding, the packed Cubic matrix occupies

\begin{equation}
T_{\rm W}=NK\left(\frac{B}{8}+\frac{8}{G}\right)\ \text{bytes}.
\label{eq:cubic-weight-bytes}
\end{equation}

The first term is the packed payload. The second is one FP32 scale and
two FP16 shape coefficients per group. Relative to a two-byte BF16
matrix, the ideal resident-weight compression factor is therefore

\begin{equation}
C_{\rm BF16/Cubic}
=\frac{2}{B/8+8/G}.
\label{eq:cubic-bf16-compression}
\end{equation}

\begingroup
\footnotesize
\renewcommand{\arraystretch}{0.86}
\begin{center}
\begin{tabular*}{0.88\linewidth}{@{\extracolsep{\fill}}crrrr@{}}
\toprule
$B$ & \shortstack{G256\\bytes/weight $\downarrow$} &
\shortstack{G256\\BF16/Cubic $\uparrow$} &
\shortstack{G512\\bytes/weight $\downarrow$} &
\shortstack{G512\\BF16/Cubic $\uparrow$} \\
\midrule
1 & 0.15625 & 12.80$\times$ & 0.140625 & 14.22$\times$ \\
2 & 0.28125 & 7.11$\times$ & 0.265625 & 7.53$\times$ \\
3 & 0.40625 & 4.92$\times$ & 0.390625 & 5.12$\times$ \\
4 & 0.53125 & 3.76$\times$ & 0.515625 & 3.88$\times$ \\
5 & 0.65625 & 3.05$\times$ & 0.640625 & 3.12$\times$ \\
6 & 0.78125 & 2.56$\times$ & 0.765625 & 2.61$\times$ \\
7 & 0.90625 & 2.21$\times$ & 0.890625 & 2.25$\times$ \\
8 & 1.03125 & 1.94$\times$ & 1.015625 & 1.97$\times$ \\
\bottomrule
\end{tabular*}
\end{center}
\endgroup

These are representation ratios, not measured kernel speedups. A
tile-local kernel reads the compressed payload and metadata without
writing a complete BF16 weight tensor, but it must generate or look up
Cubic levels and perform the same number of logical dot-product terms as
an uncompressed matrix multiplication. The leading work therefore
remains \(O(MNK)\).

\begin{longtable}[]{@{}lrr@{}}
\toprule
\begin{minipage}[b]{0.25\columnwidth}\raggedright
Component\strut
\end{minipage} & \begin{minipage}[b]{0.33\columnwidth}\raggedleft
Leading work\strut
\end{minipage} & \begin{minipage}[b]{0.33\columnwidth}\raggedleft
Ideal full-matrix traffic or state\strut
\end{minipage}\tabularnewline
\midrule
\endhead
\begin{minipage}[t]{0.25\columnwidth}\raggedright
Packed weights and metadata\strut
\end{minipage} & \begin{minipage}[t]{0.33\columnwidth}\raggedleft
Decode/lookup per visited tile\strut
\end{minipage} & \begin{minipage}[t]{0.33\columnwidth}\raggedleft
\(NK(B/8+8/G)\) bytes\strut
\end{minipage}\tabularnewline
\begin{minipage}[t]{0.25\columnwidth}\raggedright
Main dot product\strut
\end{minipage} & \begin{minipage}[t]{0.33\columnwidth}\raggedleft
\(MNK\) products and accumulations\strut
\end{minipage} & \begin{minipage}[t]{0.33\columnwidth}\raggedleft
Output-tile dependent\strut
\end{minipage}\tabularnewline
\begin{minipage}[t]{0.25\columnwidth}\raggedright
Group rescaling\strut
\end{minipage} & \begin{minipage}[t]{0.33\columnwidth}\raggedleft
\(MNK/G\) group contributions\strut
\end{minipage} & \begin{minipage}[t]{0.33\columnwidth}\raggedleft
FP32 scale/shape metadata included above\strut
\end{minipage}\tabularnewline
\begin{minipage}[t]{0.25\columnwidth}\raggedright
Dynamic-A8 activation encoding\strut
\end{minipage} & \begin{minipage}[t]{0.33\columnwidth}\raggedleft
\(MK\) reduction/rounding work\strut
\end{minipage} & \begin{minipage}[t]{0.33\columnwidth}\raggedleft
BF16 source plus an \(MK\)-byte INT8 carrier\strut
\end{minipage}\tabularnewline
\begin{minipage}[t]{0.25\columnwidth}\raggedright
Dynamic-A8 activation scales\strut
\end{minipage} & \begin{minipage}[t]{0.33\columnwidth}\raggedleft
\(M\) values for per-token A8\strut
\end{minipage} & \begin{minipage}[t]{0.33\columnwidth}\raggedleft
\(4M\) bytes under the operator semantics\strut
\end{minipage}\tabularnewline
\bottomrule
\end{longtable}

The resident Dynamic-A8 operand is approximately \(MK+4M\) bytes rather
than \(2MK\) bytes for BF16, but producing it also reads the BF16 source
and writes the carrier. Consequently, activation compression becomes a
traffic reduction only when the carrier is fused with, or reused by,
enough downstream work. A16 and Dynamic-A8 also have different constant
factors: A16 reconstructs levels into floating-point dot operands,
whereas Dynamic-A8 maps them to INT8 carriers, uses integer dot products
with INT32 partial sums, and performs FP32 group rescaling. Direct maps,
Horner evaluation, and shared-memory lookup tables alter the decode
constant but not the asymptotic complexity. This is why latency must be
measured as a function of \((M,N,K,B,G)\) rather than inferred from
storage width alone.

\hypertarget{h200-kernel-crossover-model-dtype-versus-dynamic-a8}{%
\subsection{H200 kernel crossover: model dtype versus Dynamic
A8}\label{h200-kernel-crossover-model-dtype-versus-dynamic-a8}}

The kernel experiment compares the model-dtype path, labeled A16, with
the operator-local Dynamic-A8 path on two large linear shapes at G512:
\((N,K)=(3584,3072)\) and \((6144,3584)\). These dimensions are drawn
from actual projection shapes in a widely used, large-scale open model
and serve here as representative large-model workloads; neither the
kernels nor their dispatch rules are specialized to that model. Both
paths consume the same packed Cubic representation and produce the same
output dtype. The ratio below is \(t_{\rm A16}/t_{\rm A8}\), so values
above one favor Dynamic A8. Measurements are available for
representative W2, W5, and W8 cases and do not constitute a complete
W1--W8 performance sweep.

\begin{longtable}[]{@{}rrrrr@{}}
\toprule
Weight width & \(M=1\) & \(M=16\) & \(M=64\) & \(M=256\)\tabularnewline
\midrule
\endhead
W2 & 0.40--0.59\(\times\) & 1.00--1.60\(\times\) & 1.74--2.03\(\times\)
& 2.21--2.23\(\times\)\tabularnewline
W5 & 0.78--0.87\(\times\) & 1.62--2.60\(\times\) & 2.94--3.16\(\times\)
& 3.35--3.93\(\times\)\tabularnewline
W8 & 0.35--0.42\(\times\) & 1.62--2.40\(\times\) & 2.37--4.19\(\times\)
& 4.31--4.46\(\times\)\tabularnewline
\bottomrule
\end{longtable}

For one of those shapes, \((N,K)=(3584,3072)\), the underlying measured
latencies were as follows; each cell is A16/A8 in milliseconds.

\begingroup
\small
\begin{center}
\begin{tabular*}{0.92\linewidth}{@{\extracolsep{\fill}}lrrrr@{}}
\toprule
Width & $M=1$ & $M=16$ & $M=64$ & $M=256$ \\
\midrule
W2 & 0.0184 / 0.0461 & 0.0945 / 0.0945 & 0.1653 / 0.0950 & 0.5484 / 0.2484 \\
W5 & 0.0364 / 0.0467 & 0.2658 / 0.1642 & 0.4872 / 0.1656 & 1.9521 / 0.4965 \\
W8 & 0.0171 / 0.0489 & 0.2181 / 0.1344 & 0.3950 / 0.1670 & 1.5317 / 0.3557 \\
\bottomrule
\end{tabular*}
\end{center}
\endgroup

The measurements show a crossover rather than a universal A8 advantage.
At \(M=1\), activation encoding and integer-path setup dominate, and the
mature A16 GEMV is faster. As \(M\) grows, those fixed costs are
amortized and the integer main loop becomes favorable. The crossover
also depends on width and shape, supporting the shape- and
device-dependent selection developed in Section 5.6. The measurements
are preliminary Hopper evidence rather than a general throughput claim.

\hypertarget{persistent-activation-carriers-a-negative-result}{%
\subsection{Persistent activation carriers: a negative
result}\label{persistent-activation-carriers-a-negative-result}}

An experimental path persisted groupwise INT8 activation carriers
between adjacent MLP operators to avoid an intermediate BF16 tensor.
Although this reduces the nominal intermediate width, it did not improve
either isolated or end-to-end performance and is therefore not part of
CubicQuant's retained execution design.

On a representative W1/activation-to-W2 boundary with G512 weights and G256
activation groups, the online boundary was slower than the established
non-online Dynamic-A8 path:

\begin{longtable}[]{@{}rrrr@{}}
\toprule
\begin{minipage}[b]{0.22\columnwidth}\raggedleft
Routed rows\strut
\end{minipage} & \begin{minipage}[b]{0.22\columnwidth}\raggedleft
Non-online boundary (ms) \(\downarrow\)\strut
\end{minipage} & \begin{minipage}[b]{0.22\columnwidth}\raggedleft
Online boundary (ms) \(\downarrow\)\strut
\end{minipage} & \begin{minipage}[b]{0.22\columnwidth}\raggedleft
Non-online / online\strut
\end{minipage}\tabularnewline
\midrule
\endhead
\begin{minipage}[t]{0.22\columnwidth}\raggedleft
16\strut
\end{minipage} & \begin{minipage}[t]{0.22\columnwidth}\raggedleft
0.1092\strut
\end{minipage} & \begin{minipage}[t]{0.22\columnwidth}\raggedleft
0.1408\strut
\end{minipage} & \begin{minipage}[t]{0.22\columnwidth}\raggedleft
0.776$\times$\strut
\end{minipage}\tabularnewline
\begin{minipage}[t]{0.22\columnwidth}\raggedleft
64\strut
\end{minipage} & \begin{minipage}[t]{0.22\columnwidth}\raggedleft
0.3651\strut
\end{minipage} & \begin{minipage}[t]{0.22\columnwidth}\raggedleft
0.3945\strut
\end{minipage} & \begin{minipage}[t]{0.22\columnwidth}\raggedleft
0.926$\times$\strut
\end{minipage}\tabularnewline
\begin{minipage}[t]{0.22\columnwidth}\raggedleft
128\strut
\end{minipage} & \begin{minipage}[t]{0.22\columnwidth}\raggedleft
0.7112\strut
\end{minipage} & \begin{minipage}[t]{0.22\columnwidth}\raggedleft
0.7576\strut
\end{minipage} & \begin{minipage}[t]{0.22\columnwidth}\raggedleft
0.939$\times$\strut
\end{minipage}\tabularnewline
\bottomrule
\end{longtable}

The intended bandwidth saving was real, but the producer needed
groupwise reductions, synchronization, scale generation, and an
additional carrier write. At the tested boundary those costs exceeded
the saved BF16 traffic. This negative result narrows the contribution:
Dynamic A8 is an operator-local compute mode, not a persistent
all-integer activation graph. A future design would need
producer-consumer fusion that keeps the intermediate in registers or
shared memory rather than merely changing its global-memory dtype.

\hypertarget{interpretation-of-system-level-evidence}{%
\subsection{Interpretation of system-level
evidence}\label{interpretation-of-system-level-evidence}}

Large routed operators and complete multi-GPU serving workloads have
been executed with both activation modes, but those observations combine
Cubic kernels with attention, collective communication, scheduling, and
memory management. They cannot isolate the causal contribution of the
weight format. Accordingly, this report makes no aggregate
serving-speedup, power, or energy claim.

The isolated latency results support a narrower conclusion: CubicQuant
is directly executable from its packed representation, and Dynamic A8
has a measured crossover with model-dtype execution as row count grows.
Establishing an end-to-end advantage requires matched experiments that
hold the model, parallel decomposition, attention and KV representation,
request distribution, and hardware operating point fixed. Those controls
are necessary because the full system difference cannot be attributed to
the weight format alone.

\hypertarget{related-work}{%
\section{Related work}\label{related-work}}

\hypertarget{uniform-and-second-order-post-training-quantization}{%
\subsection{Uniform and second-order post-training
quantization}\label{uniform-and-second-order-post-training-quantization}}

GPTQ uses approximate second-order information to minimize layer-wise
error during one-shot weight quantization \cite{frantar2022}. AWQ uses
activation statistics to identify salient channels and applies an
equivalent scaling transformation while retaining a hardware-friendly
low-bit format \cite{lin2023}. SmoothQuant migrates activation outlier
difficulty into weights through an equivalent transformation to enable
W8A8 execution \cite{xiao2022}. OmniQuant learns clipping and equivalent
transformations through block-wise calibration \cite{shao2023}. These
methods motivate the distinction made above: the stored representation
and the estimator used to select it are separable. CubicQuant's present
fitter is data-free, but its codes and shape parameters can be selected
with activation-aware or second-order objectives.

\hypertarget{distribution-matched-and-adaptive-scalar-formats}{%
\subsection{Distribution-matched and adaptive scalar
formats}\label{distribution-matched-and-adaptive-scalar-formats}}

QLoRA introduced NormalFloat4, a fixed non-uniform codebook constructed
for normally distributed weights, together with double quantization for
training memory reduction \cite{dettmers2023qlora}. Subsequent analysis
showed that absmax normalization changes the blockwise source
distribution and therefore the codebook favored by a reconstruction
objective \cite{yoshida2023}. BOF4 formalizes that dependence by
optimizing a blockwise codebook for a specified source model and loss
\cite{blumenberg2025}. Standard finite floating-point formats instead
place levels through a fixed exponent--mantissa split; E4M3 and E5M2 are
representative FP8 instances \cite{micikevicius2022}.

Adaptive Block-Scaled Data Types introduce IF4, which evaluates native
FP4 and scaled INT4 for each group and records the lower-MSE choice
\cite{cook2026}. IF4 therefore adapts the local scalar grid through a
discrete choice between two fixed formats; it does not learn the
reconstruction levels or continuously deform the selected grid.

LO-BCQ clusters blocks according to their local statistics and designs a
separate quantization codebook for each cluster \cite{elangovan2025}. A
block selector then identifies the shared cluster codebook used for
reconstruction. This provides finer adaptation than a single global
grid, at the cost of a codebook library and explicit selection metadata.

AAAC learns two activation-aware scalar codebooks per layer and lets
each weight group select between them \cite{islambouli2026}. CubicQuant
makes a different locality--metadata trade-off: every group receives its
own codebook, but that codebook is restricted to a monotonic cubic
family represented by two shape coefficients. Thus AAAC offers two freer
layer-level alternatives with a one-bit group selector, whereas
CubicQuant offers continuously varying group-local shapes with explicit
parameter metadata.

SqueezeLLM combines sensitivity-based non-uniform quantization with a
sparse representation for outliers \cite{kim2023}, while SpQR isolates
sensitive outliers into a higher-precision sparse component
\cite{dettmers2023spqr}. CubicQuant instead retains a homogeneous packed
scalar stream and has no separate sparse outlier matrix.

\hypertarget{vector-and-additive-codebooks}{%
\subsection{Vector and additive
codebooks}\label{vector-and-additive-codebooks}}

AQLM represents weight vectors through sums of learned codebook entries
and optimizes layer/block output reconstruction \cite{egiazarian2024}.
QuIP\# combines randomized Hadamard incoherence processing with lattice
codebooks \cite{tseng2024}. Such vector representations can express
correlations unavailable to a scalar curve and have shown strong
extreme-compression accuracy. CubicQuant instead restricts itself to a
scalar monotonic codebook to retain simple bit packing, compact
metadata, and a direct tile-local decode path. The approaches therefore
occupy different points in the accuracy, fitting-cost, metadata, and
kernel-complexity space.

\hypertarget{packed-low-bit-gpu-execution}{%
\subsection{Packed low-bit GPU
execution}\label{packed-low-bit-gpu-execution}}

LUT-GEMM replaces repeated scalar dequantization with lookup-based
products for sub-four-bit weight-only inference \cite{park2022}. FLUTE
develops this line for non-uniform LUT formats and explicitly addresses
irregular widths such as W3 through offline restructuring, vectorized
lookup, and fused matrix multiplication \cite{guo2024}. MARLIN instead
targets mixed-precision packed weights and shows that the balance
between memory traffic, dequantization, and matrix work changes with
batch size \cite{frantar2024marlin}. These systems motivate CubicQuant's
tile-local reconstruction and shape-dependent tactic selection, but do
not directly execute its parametric level function.

\hypertarget{dynamic-weight-activation-execution}{%
\subsection{Dynamic weight-activation
execution}\label{dynamic-weight-activation-execution}}

LLM.int8() exposes the importance of activation outliers and uses a
mixed-precision decomposition around an INT8 matrix product
\cite{dettmers2022llmint8}. ZeroQuant and SmoothQuant demonstrate that
backend integration, equivalent transformations, and fusion are central
to realizing W8A8 benefits \cite{yao2022,xiao2022}. QServe extends the
systems argument to W4A8 serving and identifies dequantization overhead
as a first-order concern at larger batches \cite{lin2024qserve}.
CubicQuant's Dynamic-A8 path follows the same principle---temporary
activation quantization and weight reconstruction must be consumed
inside or adjacent to the matrix kernel---but differs in how compressed
weights are mapped to the INT8 operand. Its carrier rounding is exposed
in the offline objective rather than treated as an invisible
implementation detail.

\hypertarget{limitations}{%
\section{Limitations}\label{limitations}}

\textbf{Downstream task quality remains unmeasured.} Distributional
distortion and finite-group NRMSE characterize scalar reconstruction;
they do not establish perplexity, reasoning, instruction following,
long-context behavior, or task accuracy.

\textbf{Finite-group numerical coverage is incomplete.} The reported
experiment contains one group size and omits W1 and W7. It therefore
demonstrates the observed distributional trend but cannot identify the
best group size or provide a complete 1--8-bit comparison.

\textbf{The systems evidence does not establish a general speedup.} The
H200 measurements cover representative widths and shapes rather than a
matched W1--W8 sweep across GPU generations. They establish a
workload-dependent crossover, not a universal ordering between
model-dtype and Dynamic-A8 execution.

\textbf{CubicQuant is not a native Tensor Core type.} Packed codes must
be reconstructed into model-dtype values or an INT8 carrier. Any
bandwidth benefit therefore depends on coupling reconstruction to
consumption and avoiding full-matrix expansion.

\textbf{Metadata and locality trade against one another.} Scale and
shape parameters contribute \(64/G\) bits per weight. Smaller groups
increase local adaptability but also metadata cost; larger groups
amortize metadata while forcing one curve to describe a broader tensor
region.

\textbf{Weight-only MSE is not universally optimal.} The reference
estimator gives no special weight to directions emphasized by real
activations. Covariance- or Hessian-weighted fitting may preserve
operator behavior more faithfully, but requires representative
information beyond the static weights.

\textbf{Physical performance validation is Hopper-centric.} Native
compilation targets several architectures, but the latency evidence is
drawn from H200. Compilation coverage does not substitute for numerical
and performance evaluation on Ampere, Ada, and Blackwell devices.

\textbf{Persistent activations and KV-cache formats are outside the
retained scope.} The measured persistent-activation design was
unfavorable, and attention-state quantization poses a separate numerical
and architectural problem.

\hypertarget{conclusion}{%
\section{Conclusion}\label{conclusion}}

CubicQuant separates two properties that fixed scalar formats bind
together: the payload remains a regular integer bitstream, while
group-local parameters adapt the values represented by its magnitude
codes. The normalized cubic guarantees exact zero and endpoints,
preserves ordering through an explicit monotonicity constraint, and
contains symmetric uniform integer quantization as an exact special
case. Its additional flexibility costs two shape coefficients per group,
giving a transparent effective width of \(B+64/G\) bits per weight.

The distributional analysis explains where that flexibility is useful.
The linear grid is already population-optimal for a Uniform source,
whereas Gaussian and Laplace distributions benefit from moving interior
levels toward their high-density center while retaining distant
endpoints. Across W4--W8, the optimized cubic family recovers most of
the gap between uniform integers and a free Lloyd--Max codebook for the
two unbounded reference distributions. Finite groups add a second
adaptation effect because each realized group may depart from its
population law.

The same representation supports two execution semantics. Model-dtype
kernels consume continuous Cubic levels, while Dynamic-A8 kernels
consume the deterministic INT8 projection of those levels. Incorporating
both reconstructions into the fitting objective allows one serialized
weight format to serve either path. Tile-local reconstruction preserves
compressed weight traffic, and empirical kernel selection accommodates
the fact that narrow GEMV and denser row regimes favor different
realizations.

The present evidence establishes a parametric scalar format with
explicit distortion theory, finite-group reconstruction gains, and
direct packed GPU execution. It does not yet establish downstream model
quality or a universal end-to-end speedup. Those questions require
broader numerical, architectural, and model-level evaluation, but they
do not alter the central result: a compact parametric curve can recover
much of the adaptability of a free codebook without abandoning a regular
scalar bitstream.

\appendix

\begingroup
\small

\hypertarget{normative-format-properties}{%
\section{Normative format
properties}\label{normative-format-properties}}

\begin{enumerate}
\def\labelenumi{\arabic{enumi}.}
\tightlist
\item
  \(B\in\{1,\ldots,8\}\) and \(GB\) is byte-aligned.
\item
  W1 uses binary \(-1/+1\) codes; W2--W8 reserve the most-negative
  signed pattern and decode it to zero.
\item
  The scale \(s\) is FP32 and positive for nonzero groups; \(a\) and
  \(b\) are FP16.
\item
  The coefficient \(c=1-a-b\) is derived and is not stored.
\item
  Every accepted curve satisfies \(q(0)=0\), \(q(1)=1\), and strict
  monotonicity.
\item
  Packing order and logical tensor shape determine all tail decoding.
\item
  Runtime A8 carrier generation uses round-to-nearest-even, matching the
  reference framework behavior.
\end{enumerate}

\hypertarget{reference-decoding-pseudocode}{%
\section{Reference decoding
pseudocode}\label{reference-decoding-pseudocode}}

\begin{Shaded}
\begin{Highlighting}[]
\KeywordTok{def}\NormalTok{ cubic\_levels(bits, scale, a, b):}
    \ControlFlowTok{if}\NormalTok{ bits }\OperatorTok{==} \DecValTok{1}\NormalTok{:}
        \ControlFlowTok{return}\NormalTok{ (}\OperatorTok{{-}}\NormalTok{scale, scale)}
\NormalTok{    maximum }\OperatorTok{=}\NormalTok{ (}\DecValTok{1} \OperatorTok{\textless{}\textless{}}\NormalTok{ (bits }\OperatorTok{{-}} \DecValTok{1}\NormalTok{)) }\OperatorTok{{-}} \DecValTok{1}
\NormalTok{    c }\OperatorTok{=} \FloatTok{1.0} \OperatorTok{{-}}\NormalTok{ a }\OperatorTok{{-}}\NormalTok{ b}
\NormalTok{    levels }\OperatorTok{=}\NormalTok{ []}
    \ControlFlowTok{for}\NormalTok{ index }\KeywordTok{in} \BuiltInTok{range}\NormalTok{(maximum }\OperatorTok{+} \DecValTok{1}\NormalTok{):}
\NormalTok{        t }\OperatorTok{=}\NormalTok{ index }\OperatorTok{/}\NormalTok{ maximum}
\NormalTok{        q }\OperatorTok{=}\NormalTok{ t }\OperatorTok{*}\NormalTok{ (a }\OperatorTok{+}\NormalTok{ t }\OperatorTok{*}\NormalTok{ (b }\OperatorTok{+}\NormalTok{ t }\OperatorTok{*}\NormalTok{ c))}
\NormalTok{        levels.append(scale }\OperatorTok{*}\NormalTok{ q)}
    \ControlFlowTok{return}\NormalTok{ levels}

\KeywordTok{def}\NormalTok{ decode(code, levels):}
    \ControlFlowTok{return}\NormalTok{ sign(code) }\OperatorTok{*}\NormalTok{ levels[}\BuiltInTok{abs}\NormalTok{(code)]}
\end{Highlighting}
\end{Shaded}

\hypertarget{experimental-reporting-variables}{%
\section{Experimental reporting
variables}\label{experimental-reporting-variables}}

Reproducing the numerical experiments requires the random seed, source
distribution, sample count, group size, payload width, parameter
precisions, clipping search, baseline scale optimization, and
finite-format enumeration rule. Both the objective used for candidate
selection and the statistic used for reporting must be stated
explicitly.

Reproducing kernel measurements additionally requires the GPU
architecture and operating policy, driver and CUDA versions, compiler
stack, tensor shape, activation mode, kernel family, correctness
tolerance, repetition count, and reported latency statistic.
System-level measurements require all non-Cubic components that can
affect the observation---including attention, KV state, parallel
decomposition, request distribution, and concurrency---to be held fixed
or reported separately.

\endgroup

\end{document}